\documentclass[10pt,journal,compsoc]{IEEEtran}
\ifCLASSOPTIONcompsoc
  \usepackage[nocompress]{cite}
\else
  \usepackage{cite}
\fi
\ifCLASSINFOpdf
\else
\fi

\usepackage{threeparttable}

\usepackage{xcolor}
\usepackage{tikz}
\definecolor{color1}{RGB}{93,150,72}
\definecolor{color2}{RGB}{0,143,190}
\definecolor{color3}{RGB}{181,23,0}

\usepackage[caption=false,font=footnotesize]{subfig} 

\usepackage[linesnumbered,ruled,vlined]{algorithm2e}
\usepackage{graphicx}
\usepackage{arydshln} 
\usepackage{booktabs}
\usepackage{multirow}
\usepackage{pifont}
\usepackage{color}
\usepackage{amssymb}
\usepackage{amsthm}
\usepackage{graphicx}
\usepackage{color}
\usepackage{dsfont}
\usepackage{algorithmicx}
\usepackage{algpseudocode}
\usepackage{tabularx}
\usepackage{multirow}
\usepackage{amsthm,amsmath,amssymb} 
\usepackage{mathrsfs}
\usepackage{array}
\usepackage{booktabs}  %
\usepackage{slashed}
\usepackage{scalerel}
\usepackage[top=0.4in,bottom=0.4in,left=0.4in,textwidth=7.7in]{geometry}
\usepackage{xspace}
\makeatletter 
\DeclareRobustCommand\onedot{\futurelet\@let@token\@onedot}
\def\@onedot{\ifx\@let@token.\else.\null\fi\xspace}

\def\eg{\emph{e.g}\onedot} 
\def\ie{\emph{i.e}\onedot} 
 
\def\etc{\emph{etc}\onedot} 
 
\def\etal{\emph{et al}\onedot}
\makeatother

\usepackage[pagebackref=false,breaklinks=true,linkcolor=red,anchorcolor=blue, citecolor=green,colorlinks,bookmarks=false]{hyperref}
\begin{document}

\title{A Causal Adjustment Module for Debiasing \\ Scene Graph Generation}

\author{
Li Liu, Shuzhou Sun, Shuaifeng Zhi, Fan Shi, Zhen Liu, Janne Heikkilä, Yongxiang Liu 
\IEEEcompsocitemizethanks{
\IEEEcompsocthanksitem This work was supported by the National Key Research and Development Program of China No. 2021YFB3100800, the National Natural Science Foundation of China under Grant 62376283 and 62201603, the Science and Technology Innovation Program of Hunan Province under Grant 2021RC3069, and the Key Stone grant (JS2023-03) of the National University of Defense Technology (NUDT), The the CPSF (No. 2023TQ0088, No. GZC20233539), and the Research Program of NUDT (No. ZK22-04).
\IEEEcompsocthanksitem Li Liu (liuli\_nudt@nudt.edu.cn), Shuaifeng Zhi, Zhen Liu and Yongxiang Liu are with the College of Electronic Science and Technology, NUDT, Changsha, Hunan, China. Fan Shi is with the College of Electronic Engineering, NUDT, Hefei, Hunan, China.
\protect
\IEEEcompsocthanksitem Yongxiang Liu and Zhen Liu are the corresponding authors.
\IEEEcompsocthanksitem Shuzhou Sun is with the with the Department of Computer Science \& Technology, Tsinghua University, Beijing 100190, China, and also with the Center for Machine Vision and Signal Analysis (CMVS), University of Oulu, 90570 Oulu, Finland. Janne Heikkila is with the Center for Machine Vision and Signal Analysis (CMVS), University of Oulu, 90570 Oulu, Finland.
}
}

\markboth{IEEE Transactions on Pattern Analysis and Machine Intelligence}%
{Sun \MakeLowercase{\textit{et al.}}: USGG}

\IEEEtitleabstractindextext{%
\begin{abstract}
While recent debiasing methods for Scene Graph Generation (SGG) have shown impressive performance, these efforts often attribute model bias solely to the long-tail distribution of relationships, overlooking the more profound causes stemming from skewed object and object pair distributions. In this paper, we employ causal inference techniques to model the causality among these observed skewed distributions. Our insight lies in the ability of causal inference to capture the unobservable causal effects between complex distributions, which is crucial for tracing the roots of model bias. Specifically, we introduce the Mediator-based Causal Chain Model (MCCM), which, in addition to modeling causality among objects, object pairs, and relationships, incorporates mediator variables, \ie, cooccurrence distribution, for complementing the causality. Following this, we propose the Causal Adjustment Module (CAModule) to estimate the modeled causal structure, using variables from MCCM as inputs to produce a set of adjustment factors aimed at correcting biased model predictions. Moreover, our method enables the composition of zero-shot relationships, thereby enhancing the model's ability to recognize such relationships. Experiments conducted across various SGG backbones and popular benchmarks demonstrate that CAModule achieves state-of-the-art mean recall rates, with significant improvements also observed on the challenging zero-shot recall rate metric.
\end{abstract}

}

\maketitle

\IEEEdisplaynontitleabstractindextext
\IEEEpeerreviewmaketitle

\IEEEraisesectionheading{\section{Introduction}\label{sec:introduction}}

Scene Graph Generation (SGG) \cite{ImageRetrieval,Neuralmotifs,SGGSurvey} aims at creating a semantic and  structured representation of a scene in the form of a graph that describes objects (\eg, ``\emph{man}'', ``\emph{chair}'') and their attributes (\eg, ``\emph{chair is white}''), as well as relationships between paired objects (``\emph{man sitting on chair}''), which is typically formulated as a set of \textit {$<$\emph{subject}, \emph{predicate}, \emph{object}$>$} triplets (\eg, $<$\emph{man}, \emph{sitting on}, \emph{chair}$>$).  By transforming images into structured, detailed semantics, scene graphs facilitate a higher-level understanding of visual content,  enhancing the ability to understand and reason about visual scenes.
 SGG is crucial for many advanced computer vision applications, including image generation \cite{imagegeneration1,imagegeneration2}, image and video captioning \cite{imagecaptioning1,imagecaptioning2}, cross-modal retrieval~\cite{ImageRetrieval,retrieval},
visual question answering \cite{VQA1,VQA2}, and 3D scene understanding \cite{3d-scene-generation1,3d-scene-generation2}. 



\begin{figure*}
    \footnotesize\centering
    \centerline{\includegraphics[width=1\linewidth]{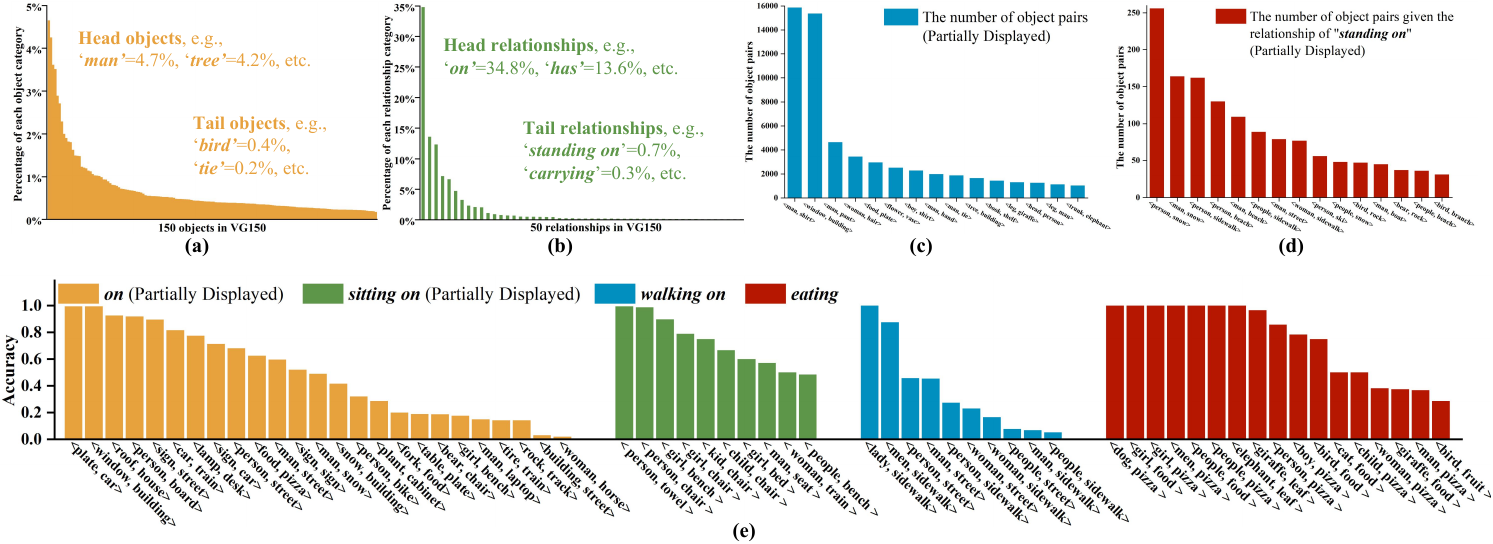}}
        \vspace{-0.2cm}
        \caption{The motivations of our proposed CAModule. (a): Distribution of 150 object categories. (b): Distribution of 50 relationship categories. (c): Distribution of object pairs. Theoretically, there could be 150×150 possible object pair combinations. Although the actual occurrence of object pairs is sparse, only a small number of pairs is displayed. (d): Distribution of object pairs within triplets formed by the ``standing on'' relationship. (e): Accuracy for different triplets across various relationships. Here, for a more intuitive display of the model's performance across diverse triplets, accuracy rather than recall rate is computed. Experiments in this figure are conducted on the VG150 \cite{VG150} dataset.}
    \label{motivation}
\end{figure*}

While SGG has witnessed significant progress, however, it still faces several challenges, among which the problem of longtail distribution in visual relationship recognition is a key one. A long-tailed distribution in visual relations is inevitable due to the wide variety of visual relationships and the complex nature of realistic scenes. More specifically, the predicates in visual relationships are longtailed, and common head predicates (\eg,
``\emph{on}'',  ``\emph{has}'', ``\emph{in}'') appear frequently, while others are rare. As illustrated in Fig. \ref{motivation} (b), the head relationship ``\textit{on}'' accounts for $34.8\%$, while the tail relationship ``\textit{carrying}'' comprises only $0.3\%$. The actual long-tailed
distribution of relationships existing in SGG datasets can have a large influence on the accuracy and
completeness of the generated scene graph. In addition, the more accurate, tail relationship expressions are usually predicted as less informative, overly
general ones (\eg, predicting \textit {$<$\emph{people}, \emph{sitting on}, \emph{chair}$>$} as \textit {$<$\emph{people}, \emph{on}, \emph{chair}$>$}). The tendency of the model to predict a trivial and general relationship over a more precise one is referred to as Biased SGG. This bias problem makes it challenging to learn models that can accurately recognize less common relationships.  The community has realized that there is a pressing need to develop debiasing methods to address the long-tail problem. Currently, such methods include the following categories:
\begin{itemize}
  \item \textit {Resampling methods}  \cite{SegG,TransRwt,DT2,TBE,CFA} that oversample tail relationships or undersample head ones;
  \item \textit {Reweighting methods} \cite{Cogtree,FGPL,PPDL,GCL,ML-MWN,EBMloss,DKBL,gao2023informative} that adjust the influence of different relationship categories on model training to enhance learning of tail relationships;
  \item \textit {Adjustment methods} \cite{DLFE,TDE,logit-adjustment-SGG,logitadjustment} that recalibrate biased model outputs, typically by increasing weights for tail categories and decreasing them for head classes;
  \item  \textit {Hybrid methods} \cite{HML,CAME,RTPB,NICEST,BPLSA} that combine some or all of the aforementioned strategies.
\end{itemize}

Although these methods represent progress in debiasing, they have clear limitations. For the resampling methods, oversampling can lead to model overfitting, while undersampling may cause underfitting. In attempts to improve the detection of tail relationships, reweighting and adjustment methods may substantially hinder the identification of head categories \cite{add_longtail_2,add_longtail_3}.  More crucially, these debiasing methods primarily attribute model bias to the long-tail problem of relationships, overlooking the fact that the skewed relationship distributions stem from the similarly long-tailed issues of both objects and object pairs. The underlying rationale is that relationships are generated by object pairs, and the potential relationships that can be formed by certain object pairs are relatively fixed, such as \textit {$<$man, shirt$>$} likely forming the relationship ``\textit{wears}'' instead of ``\textit{eating}''. Hence, if object pairs follow a long-tail distribution, the resultant relationship distribution will also be skewed. Furthermore, the distribution of objects dictates the distribution of object pairs since the probability of forming an object pair increases with the abundance of each object. In fact, for the SGG task, both object and object pair distributions face severe long-tail distribution issues, as shown in Fig. \ref{motivation} (a) and (c).


Based on these observations, this paper proposes the use of causal inference to capture the causal relationships among object distribution $\mathcal{O}$, object pair distribution $\mathcal{P}$, and relationship distribution $\mathcal{R}$. Causal inference is a procedure that employs observed data and plausible assumptions to estimate causal effects that cannot be directly observed \cite{pearl2009causality,pearlCBM,PearlPCH}. Our primary insight is to thoroughly trace and rectify the biases in SGG models caused by these observed skewed distributions. As an initial step in integrating causal inference techniques within the SGG task, we first causally model the observed skewed distributions. Interestingly, our analysis of these distributions \ie, $\mathcal{O}$ affecting $\mathcal{P}$, which in turn influences $\mathcal{R}$, aligns well with the typical two-stage framework of SGG task. Therefore, we naturally model these skewed distributions as a chained structure, the Causal Chain Model (CCM), \ie, $\mathcal{O} \rightarrow \mathcal{P} \rightarrow \mathcal{R}$. While intuitive, we believe that CCM still does not fully capture the causality among these causal variables. This is particularly true for $\mathcal{O} \rightarrow \mathcal{P}$, which underscores the impact of object distribution on object pair distribution, implying the more numerous two objects are, the more pairs they should form. However, we argue that this overlooks the varying likelihood of any two objects forming a reasonable pair. Our statistical analysis in Fig. \ref{motivation} (a) and (c) supports this; for instance, ``\textit{man}'' and ``\textit{tree}'' rank among the most numerous, yet \textit {$<$man, tree$>$} is not among the most frequent object pairs. We attribute this to the omission of cooccurrence, \ie, the probability of any two objects appearing together in the same scene. Inspired by this observation, we propose an advancement over the CCM structure, the Mediator based Causal Chain Model (MCCM). MCCM introduces a mediator variable, the co-occurrence distribution $\mathcal{C}$, between $\mathcal{O}$ and $\mathcal{P}$, thus optimizing $\mathcal{O} \rightarrow \mathcal{P}$ to $\mathcal{O} \rightarrow \mathcal{C} \rightarrow \mathcal{P}$. MCCM addresses the inadequacies in causal relationship modeling of CCM by incorporating the mediator variable $\mathcal{C}$, thereby enhancing the tracing of model biases.

Building on MCCM, we introduce the Causal Adjustment Module (CAModule) to estimate causal relationships among variables within this structure. Specifically, the CAModule takes as inputs the skewed distributions $\mathcal{O}$, $\mathcal{C}$, $\mathcal{P}$, $\mathcal{R}$, and outputs a set of adjustment factors that can adjust the outputs of biased SGG models. While bearing similarities to traditional logit adjustment methods used for simple classification tasks \cite{logitadjustment}, our approach is fundamentally different: it employs triplet-level adjustments, assigning an adjustment factor to each triplet, unlike traditional methods that implement relationship-level adjustments, where one factor corresponds to one relationship (please refer to Fig. \ref{adjust} to find the difference). Moreover, whereas traditional methods might utilize prior knowledge, such as the inverse of category frequencies, as adjustment factors, our method derives these factors from learning within the causal module. Owing to its fine-grained logit adjustment capabilities, our proposed CAModule achieves state-of-the-art performance on the debiasing metric mean recall rate for the SGG task.

Beyond achieving state-of-the-art performance on the mean recall rate metric, our method has surprisingly shown considerable efficacy in identifying the notably challenging zero-shot relationships. We attribute this advantage to our method's ability to compose zero-shot relationships, with further analysis available in Section \ref{sec3.4}. However, it's noteworthy that the composition units for zero-shot relationships in our approach are object pairs and relationships. This highlights a limitation: our method cannot generate zero-shot relationship from zero-shot pair, \ie, the object pair that do not appear in the training set but are present in the test set. To address this limitation, we propose optimizing the object pair distribution $\mathcal{P}$ to include as many zero-shot pairs as possible. Specifically, we suggest inferring unknown zero-shot pairs based on observed object pairs, under the principle that objects with similar attributes may form the same relationships. For instance, if \textit {$<$girl, sitting on, chair$>$} is observed, then \textit {$<$boy, sitting on, chair$>$} is also considered a plausible relationship due to the similarity in attributes between ``\textit {girl}'' and ``\textit {boy}''. To this end, we establish two inference rules, \textbf{Rule 1} and \textbf{Rule 2}, with the former emphasizing object similarity and the latter subject similarity, for detailed inference rules, refer to Section \ref{sec3.4}. Leveraging these rules, our method can identify zero-shot pairs and, consequently, compose zero-shot relationships, thereby further improving the recognition of such relationships.

In summary, the contributions of our work are three-fold:
\begin{itemize}
\item We conduct a thorough analysis of bias within SGG models, identifying not only the widely discussed long-tail distribution issue of relationships but also uncovering that skewed distributions of objects and object pairs serve as deeper underlying causes of bias.

\item Utilizing the causal inference technique, we trace the origins of bias in SGG models by modeling the skewed distributions of objects, object pairs, and relationships. Specifically, we propose the Mediator-based Causal Chain Model (MCCM), which introduces mediator variables, \ie, co-occurrence distribution, to complement the causality between objects and object pairs. To estimate MCCM, we design the Causal Adjustment Module (CAModule) that takes causal variables from MCCM as input and outputs a set of triplet-level adjustment factors capable of correcting biased model predictions. Moreover, our method enhances the model's capability to recognize zero-shot relationships, and we also propose some inference rules for object pairs to strengthen this ability further.

\item Comprehensive experiments conducted across various SGG backbones and popular benchmarks demonstrated the state-of-the-art mean recall rates achieved by the proposed CAModule. Furthermore, our method also shows substantial improvement on the challenging zero-shot recall rate metric.
\end{itemize}

\section{Related works}
\subsection{Scene Graph Generation}
\label{Related-works-SGG}
SGG aims to automatically infer the various relationships between objects in an image and organize this information into a graphical structure, known as a scene graph \cite{ImageRetrieval,SGGPAMI,tian2021mask,gao2023informative,tian2020part}. Early efforts focused on optimizing network structures to better represent the features of relationships between objects, such as those involving CNN-based \cite{CNN-SGG-1,CNN-SGG-2}, RNN/LSTM-based \cite{Neuralmotifs,VCtree}, and graph-based \cite{GNN-SGG-1,GNN-SGG-2} models. Due to the extremely skewed distribution of datasets, recent work has shifted focus to the bias issue in SGG models, which tend to predict fine-grained relationships as coarse-grained ones. Given the focus of this paper, we only review debiasing methods in SGG task, which can be broadly categorized into four groups: 

\textit {Resampling methods} \cite{SegG,TransRwt,DT2,TBE,CFA} address the severe imbalance issue in datasets by downsampling head relationships or upsampling tail categories. The sampling process can be random; however, more often, it is purposefully guided by prior knowledge, such as object attribute priors. For instance, Compositional Feature Augmentation (CFA) \cite{CFA} designed two distinct feature augmentation modules, the intrinsic and extrinsic modules, to complement the feature diversity of the original relational triplets by replacing or mixing the features of other samples, based on the similarity of object attributes. Recently, Group Collaborative Learning (GCL) \cite{GCL} reorganized the dataset by splitting the extremely unbalanced dataset into a set of relatively balanced groups.


\textit {Reweighting methods} \cite{Cogtree,FGPL,PPDL,GCL,ML-MWN,EBMloss,DKBL,gao2023informative} enhance learning of tail relationships by adjusting the influence of different categories during the model's training phase, typically by weighting the prediction loss. The weighting rules are often based on prior knowledge extracted from observed data, such as the cognitive structure in Cognition tree (CogTree) \cite{Cogtree}, predicate lattice in fine-grained Predicates Learning (FGPL) \cite{FGPL}, and relationship probability distribution in Predicate Probability Distribution based Loss (PPDL) \cite{PPDL}. Recently, Dark Knowledge Balance Learning (DKBL) \cite{DKBL} employs weights to the loss function based on dark knowledge, \ie, the predictive probability scores of negative samples, allowing the model to balance head and tail categories while maintaining overall performance. 


\textit {Adjustment methods} \cite{DLFE,TDE,logit-adjustment-SGG,logitadjustment} modify the model's outputs for debiasing predictions. For instance, Chiou \etal \cite{DLFE} treat SGG as a Learning from Positive and Unlabeled data (PU learning) task and introduces Dynamic Label Frequency Estimation (DLFE) to refine label frequency estimation. Adaptive Logit Adjustment Loss (ALA Loss) \cite{logit-adjustment-SGG} introduces an adaptive logit adjustment method, with the adjusting term comprising two complementary factors: the quantity factor and the difficulty factor. 


\textit {Hybrid methods} \cite{HML,CAME,RTPB,NICEST,BPLSA} integrate parts or all of the aforementioned strategies. For instance, Hierarchical Memory Learning (HML) \cite{HML} and Context-Aware Mixture-of-Experts (CAME) \cite{CAME} initially divide the imbalanced training data into multiple relatively balanced subsets. Subsequently, HML trains the model with coarse-grained relationships and fine-tunes it for fine-grained categories, whereas CAME proposes utilizing a mixture of experts to handle these subsets.

While debiasing research is rather active within the SGG community, these methods have their limitations. Resampling methods carry inherent risks of overfitting and underfitting, reweighting methods may result in poor and unstable performance due to the sensitivity of adjustment factors, and adjustment methods may hert performance for certain relationship categories. Inspired by traditional adjustment methods \cite{add_longtail_2,add_longtail_3,logitadjustment}, this paper introduces a fine-grained logit adjustment strategy for model debiasing. However, our approach fundamentally differs from traditional methods in several aspects: \textbf{1)} Beyond the widely studied long-tail distribution of relationships, this paper also identifies the imbalance in object and object pair distributions as a source of model bias; \textbf{2)} In modeling the causality of these distributions, we introduce a mediator variable, the co-occurrence distribution, to intervene and optimize the causal structure, an aspect not yet explored in existing debiasing efforts to our knowledge; \textbf{3)} The adjustment factors in our method are at the triplet-level, effectively preventing cross-relationship interference during the adjustment process.

\begin{figure*}
    \footnotesize\centering
    \centerline{\includegraphics[width=1\linewidth]{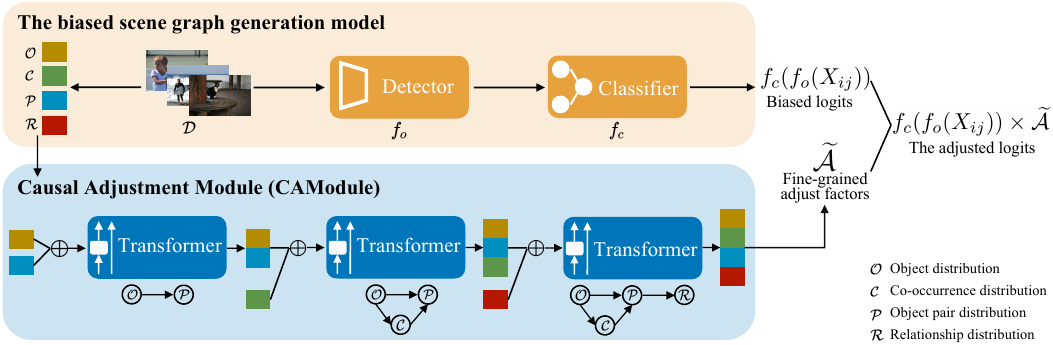}}
        \vspace{-0.2cm}
        \caption{The typical pipeline of a biased scene graph generation model and the proposed Causal Adjustment Module (CAModule) of this paper. The SGG pipeline primarily comprises an object detector and a relationship classifier, where the former detects the categories and positional information of objects within the images, while the latter classifies the relationship features of each pair of objects. CAModule takes the object distribution, co-occurrence distribution, object pair distribution, and relationship distribution as inputs and outputs a set of fine-grained adjustment factors for adjusting the logits output by the biased SGG model.}
    \label{framework}
    \vspace{-0.3cm}
\end{figure*}

\subsection{Causal Inference}
\label{Related-works-Causal-Inference}
Causal inference is a procedure utilized for analyzing and understanding the causal relationships between variables, aiming primarily to distinguish between correlation and causation. It has made profound impacts in fields such as statistics, econometrics, and epidemiology \cite{pearl2009causality,pearlCBM,PearlPCH,Bengio2021Toward,Wang,DisC}. Recently, causal inference has attracted extensive attention in deep learning community due to its potential in enhancing model transparency, interpretability, and generalizability by understanding causal relations in complex data \cite{causalGeneralization,causal_ood,causal_detection}. Some efforts have been made to mitigate biases and spurious correlations in SGG models through causal inference. For instance, Total Direct Effect (TDE) \cite{TDE} reconceptualizes the SGG task within a causal graph framework, suggesting the derivation of counterfactual causal relationships from trained graphs to eliminate bad biases in models. Two-stage Causal Modeling (TsCM) \cite{TsCM} takes long-tailed distribution and semantic confusion as biases in SGG task and proposes causal representation learning and causal calibration learning to eliminate these biases, respectively. DSDI \cite{DSDI} addresses dual imbalances in observed data by incorporating a bias-resistant loss and a causal intervention tree, thereby enhancing foreground representation and mitigating context bias. Lu \etal \cite{lu2023prior} apply common-sense-based causal reasoning strategies to SGG models, utilizing interventions and counterfactual reasoning to alleviate spurious correlations between context information and predicate labels.

Building on similar insights, this paper proposes the use of causal inference modeling for distributions derived from observed data, and introduces a lightweight module to capture the modeled causality. Our approach finely adjusts the the biased outputs in SGG models. Compared to the aforementioned causal inference-based efforts, our method underscores two distinct aspects: \textbf{1)} Our analysis of model bias is more thorough. Beyond relationship distribution, we also extract and causally model object distribution, co-occurrence distribution, and object pair distribution from observational data to analyze the causes of model bias, where the co-occurrence distribution is an aspect not yet explored by existing works. \textbf{2)} Our method not only achieves the best performance on the main debiasing metric mR@K but also significantly improves the challenging zero-shot relationship recognition metric zR@K, a feat not accomplished by other similar works. This improvement is due to our method's involvement in object pair and relationship combinations, enabling the composition of partial zero-shot relationships and thereby enhancing recognition performance for such relationships.

\section{Methods}

\subsection{Overview}
\label{Overview}

Scene graph generation is a pivotal task that bridges computer vision and natural language processing by identifying and organizing various objects and their relationship within an image into a graphical structure, known as a scene graph  \cite{ImageRetrieval,SGGSurvey,SGGPAMI}. As illustrated in Fig. \ref{framework}, the standard SGG model typically include an object detector $f_o$ and a relationship classifier $f_c$ \cite{Neuralmotifs,VCtree,TDE}. The former detects object information, whereas the latter classifies the inter-object relationships. Formally, denote the observed data as $\mathcal{D}$, encompassing $N_{o}$ object categories $\mathbf{o} = \{o_i\}_{i=1}^{N_{o}}$ and $N_{r}$ relationship classes $\mathbf{r} = \{r_i\}_{i=1}^{N_{r}}$. In this context, the probability that a relationship instance, composed of objects from category $o_i$ and category $o_j$, denoted as $X_{i j}$, belongs to the $k$-th category of relationship is:
\begin{equation}
P(Y_{i j k}=k)= \sigma( f_c(f_o(X_{i j})) ),
\end{equation}
where $Y_{i j k}$ represents the triplet composed of the $i$-th and $j$-th object classes and the $k$-th relationship category; $X_{i j}$ denotes the original features of $Y_{i j k}$ in the image. $\sigma$ is a function that transforms $f_c$' output into probabilities, typically the sigmoid function. 

In earlier years, SGG research primarily focused on designing efficient network models to capture relationship features between objects, such as CNN-based \cite{CNN-SGG-1,CNN-SGG-2}, RNN/LSTM-based \cite{Neuralmotifs,VCtree}, and graph-based \cite{GNN-SGG-1,GNN-SGG-2} models. However, in recent years, the emphasis has shifted towards model bias, a phenomenon where fine-grained tail relationships are predicted as coarse-grained head ones \cite{TDE,VCtree}. The majority of studies attribute this bias to the long-tail distribution of relationships, as illustrated in Fig. \ref{motivation} (b). Naturally, we are curious about how the long-tail distribution affects model bias. As an example, for a specific relationship instance, how does the model's prediction change when it is considered as a head category versus a tail one? Although it may seem counterintuitive, counterfactual reasoning can capture these variations in model predictions:
\begin{equation}
\begin{aligned}
Y_{i j k}^{d o(x)}=&\int_x[E(Y_{i j k} \mid \operatorname{do}(X_{i j}=x)) \\
& -E(Y_{i j k} \mid \operatorname{do}(X_{i j}=x_0))] p(x) d x ,
\label{counterfactual}
\end{aligned}
\end{equation}
where $\operatorname{do}(X_{i j}=x)$ and $\operatorname{do}(X_{i j}=x_0)$ respectively represent $X_{i j}$ as a head and a tail instance after causal intervention. In this manner, the Total Direct Effect (TDE) simulates counterfactual reasoning by purposefully erasing parts of an image to achieve debiasing predictions:
\begin{equation}
P(Y_{i j k}=k)= \sigma( f_c(f_o(X_{i j})) - f_c(f_o(\widetilde{X}_{i j}) )) ,
\end{equation}
where $X_{i j}$ and $\widetilde{X}_{i j}$ respectively denote instances before and after the treatment. A clear limitation is that for each instance, TDE requires two model inferences, introducing additional computational burden.

In this paper, we introduce a counterfactual estimation method that requires only a single model inference, inspired by the logit adjustment approach. Logit adjustment, widely researched in long-tail classification tasks, enhances the performance of tail categories by adjusting the logits of model outputs \cite{logitadjustment}. Considering the long tail nature of relationships in the SGG task, this method can be directly applied to SGG:
\begin{equation}
P(Y_{i j k}=k)=\sigma(f_c(f_o(X_{i j})) \times \mathcal{A}),
\label{relation-level-adjust}
\end{equation}
\begin{equation}
\mathcal{A} = \{ a_1, a_2, \ldots, a_{N_r} \},
\end{equation}
where $\mathcal{A}$ is a set of adjustment factors, typically the inverse of the frequency of each category. For instance, $a_i$ could be the inverse of the frequency of the $i$-th relationship category, serving as the adjustment factor for this relationship. Although the logit adjustment method performs well in simple classification tasks, unfortunately, it shows almost no debiasing effect on the model in the SGG task (see Section \ref{Ablations}). We attribute this to the dual imbalance in the SGG dataset, specifically object imbalance and relationship imbalance.  This dual imbalance implies that some relationship categories, although classified as head classes, may form triplets that fall into the tail ones. For example, '\textit {on}' is a head relationship category, but the triplet  \textit {$<$bag, on, arm$>$ }  accounts only 0.037\% of the training instances.
Imbalance at the triplet-level implies that triplets composed of the same relationship can exhibit significantly varied performances. As shown in Fig. \ref{motivation} (e), although formed by the same relationship '\textit {on},' triplets \textit {$<$plate, on, car$>$ } and \textit {$<$window, on, building$>$ } have accuracies close to 100\%, while those of \textit {$<$woman, on, horse$>$ } and \textit {$<$building, on, street$>$}  are nearly 0\%. This observation suggests that, even for triplets composed of the same relationship, we should apply distinct adjustment factors, rather than a single adjustment factor per category as done in traditional methods. Inspired by this observation and benefiting from the concept of Average Treatment Effect (ATE) \cite{ATE} in causal inference, we propose a triplet-level intervention approach:
\begin{equation}
\operatorname{ATE}(k)=\frac{1}{\Omega^2} \sum_{i, j \in \Omega} \operatorname{Effect}(i, j, k),
\label{ATE}
\end{equation}
where $\Omega$ denotes the set containing all possible object categories in the dataset, hence, $\Omega^2$ represents all possible object combinations in observed data $\mathcal{D}$. $i$, $j$, $k$ respectively represent the indices of object, subject, and relationship in a triplet relationship, with $\operatorname{Effect}(i, j, k)$ thus considering both relationship and object information. ATE is a core concept in causal inference, used to assess the average effect of a treatment or intervention across a population \cite{pearl2009causality,PearlPCH}. Building on $\operatorname{ATE}(k)$, we further develop the Causal Adjustment Module (CAModule) to obtain the triplet-level logit adjustment factors $\widetilde{\mathcal{A}}$:
\begin{equation}
\widetilde{\mathcal{A}} = \{ \widetilde{\mathcal{A}}_1, \widetilde{\mathcal{A}}_2, \ldots, \widetilde{\mathcal{A}}_{N_r} \},
\end{equation}
\begin{equation}
\widetilde{\mathcal{A}}_k = \begin{pmatrix}
\widetilde{a}_{1k1} & \widetilde{a}_{1k2} & \cdots & \widetilde{a}_{1kN_{o}} \\
\widetilde{a}_{2k1} & \widetilde{a}_{2k2} & \cdots & \widetilde{a}_{2kN_{o}} \\
\vdots  & \vdots  & \ddots & \vdots  \\
\widetilde{a}_{N_{o}k1} & \widetilde{a}_{N_{o}k2} & \cdots & \widetilde{a}_{N_{o}kN_{o}} 
\end{pmatrix},
\end{equation}
where $\widetilde{a}_{ikj}$ denotes the adjustment factor for instances composed of the $i$-th and $j$-th objects and the $k$-th relationship. Following the standard logit adjustment procedure, we use the triplet-level adjustment factors $\widetilde{\mathcal{A}}$ to modify the logits from the baseline SGG model's output, aiming for debiasing predictions:
\begin{equation}
P(Y_{i j k}=k)=\sigma(f_c(f_o(X_{i j})) \times \widetilde{\mathcal{A}}).
\label{triplet-level-adjustment}
\end{equation}

As a lightweight module, CAModule takes statistical knowledge derived from observed data $\mathcal{D}$ as its input and outputs fine-grained triplet adjustment factors $\mathcal{A}$. Notably, in addition to the object and relationship distributions widely used in existing debiasing work, we also introduce co-occurrence and object pair distributions; the former quantifies the probability of any two objects appearing together in the same scene, while the latter measures the likelihood that any two objects can constitute a reasonable pair. To explore the potential contributions of these distributions to SGG model bias, we first build a Mediator-based Causal Chain Model (MCCM) to reorganize the causality of these distributions. This causal model employs co-occurrence and object pair distributions as mediators, providing additional insights into the intrinsic pathways of the Causal Chain Model (CCM) adhered to by the typical SGG training paradigm, thereby mitigating bias induced by skewed distributions.

In Section \ref{sec3.2}, we conduct a detailed analysis on the establishment of causal structures, from traditional causal chain structure to the mediator-variable-based structure proposed by us. Subsequently, in Section \ref{sec3.3}, we elaborate on the structure of the CAModule and analyze how it captures the causal pathways within MCCM. Owing to the fine-grained adjustment capabilities of CAModule, our method achieves competitive results on key debiasing metrics mean Recall Rate (mR@K) (refer to Section \ref{results}). Additionally, to our surprise, we found a notable improvement in the challenging and underexplored metric of zero-shot Recall Rate (zR@K). Our preliminary analysis suggests that this improvement is attributed to our developed MCCM, capable of combining certain potential zero-shot relationships. This exciting discovery inspires us to further optimize the mediator variable of object pair distribution to enhance this combinatorial ability for such relationships, with detailed analysis and implementation presented in Section \ref{sec3.4}.

\begin{figure}
    \footnotesize\centering
    \centerline{\includegraphics[width=0.8\linewidth]{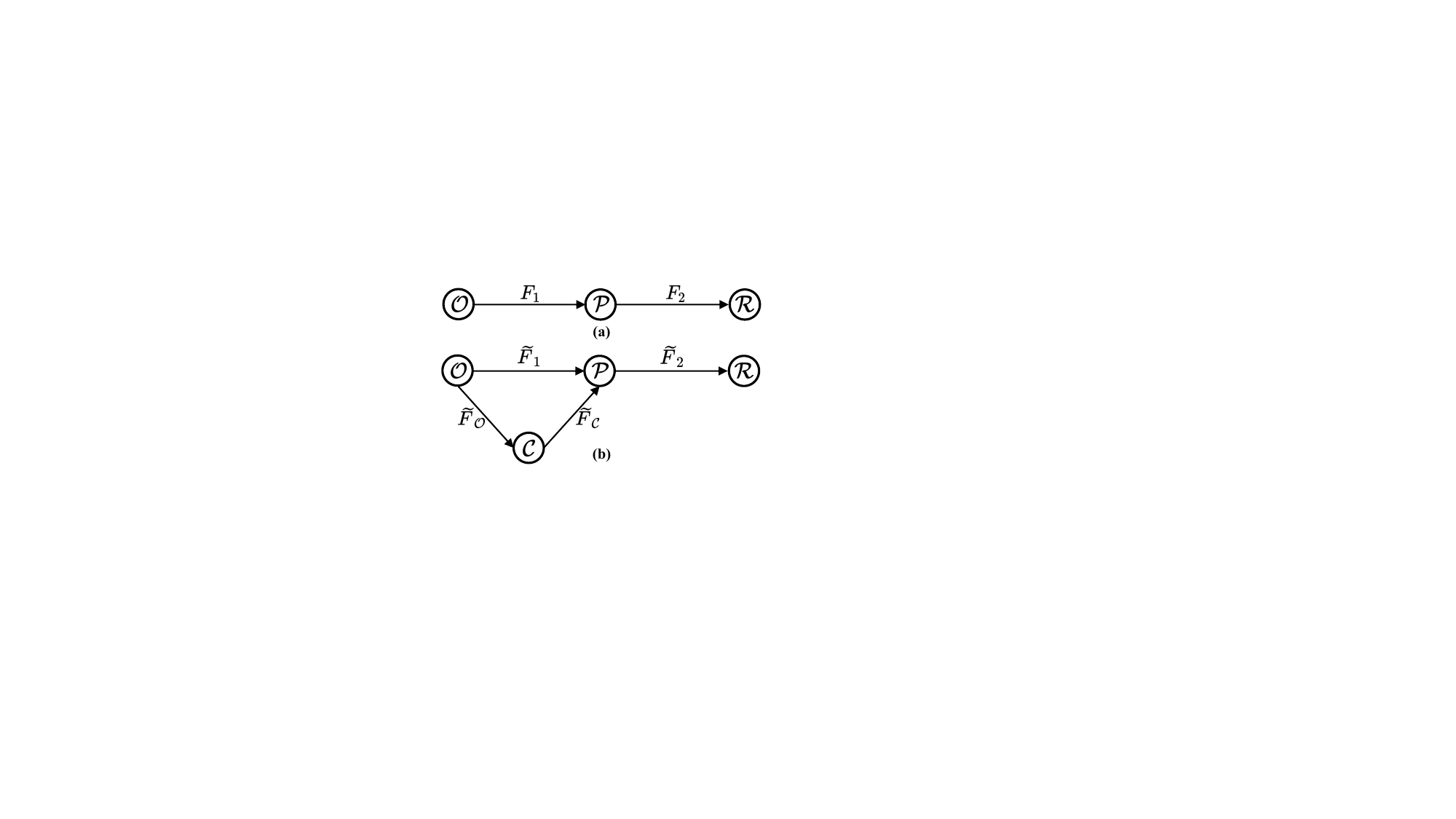}}
        \vspace{-0.2cm}
        \caption{Structural Causal Model (SCM) of typical scene graph generation framework (a) and our proposed method (b). $\mathcal{O}$, $\mathcal{C}$, $\mathcal{P}$, and $\mathcal{R}$ represent the distributions of object, co-occurrence, object pair, and relationship, respectively.}
    \label{SCM}
\end{figure}

\subsection{Modeling Structural Causal Model}
\label{sec3.2}

To understand the causality between variables for making accurate predictions and decisions, a variety of causal modeling methods have been developed, such as Causal Graphical Models (CGM), Potential Outcomes (PO), and Structural Causal Models (SCM) \cite{pearlCBM,CausalFairness,pearl2009causality,StatisticaltoCausal}. CGM graphically represents causal relationships between variables, yet its accuracy heavily depends on the correct specification of the graph's edges. PO focuses on interventions on individuals and observing the outcomes, but its capability to investigate complex causal mechanisms between variables is limited. SCM is a formalized causal modeling approach that describes the causal relationships through a set of structural equations, making it particularly suited for intervention analysis and counterfactual reasoning \cite{pearl2009causality,pearlCBM,PearlPCH}. As introduced in Section \ref{Overview}, we utilize counterfactual inference to pursue debiasing predictions; thus, in this paper, we adopt SCM for causal modeling. SCM mathematically represents causal mechanisms, including variables, structural functions, and distributions over the variables. 

As depicted in Fig. \ref{framework}, the SGG model typically comprise an object detector $f_o$ and a classifier $f_c$, where the former extracts object information, often including visual features, bounding boxes, and object categorization, while the latter pairs objects and assigns a relationship to each pair. Therefore, the training paradigm of SGG model follows a chain structure, where the detector's inputs determine the classifier's inputs, which in turn influence the final predictions. As a result, the SGG task can be modeled as a Causal Chain Model (CCM) (see Fig. \ref{SCM} (a)): 
\begin{equation}
\begin{aligned}
& \mathcal{O}=P(\mathcal{O}), \\
& \mathcal{P}=F_1(\mathcal{O}, P(\mathcal{O})), \\
& \mathcal{R}=F_2(\mathcal{P}, P(\mathcal{P})),
\end{aligned}
\end{equation}
where $\mathcal{O}$, $\mathcal{P}$, and $\mathcal{R}$ respectively represent the distribution of objects, object pairs, and relationships. $F_1$ and $F_2$ denote structural functions among causal variables. Under such a setup, we can obtain the conditional probability of the outcome variable $\mathcal{R}$:
\begin{equation}
P_{\text {CCM}}(\mathcal{R} \mid \mathcal{O}, \mathcal{P})=\sum_{o, p} P(\mathcal{R} \mid o, p) \cdot P(o \mid \mathcal{O}) \cdot P(p \mid \mathcal{P}).
\end{equation}
While causal chain models emphasize that triplets in scene graphs are composed of two objects and the relationship between them, they overlook \textbf{1)} the variability in the occurrence frequency of different object pairs, such as the count of \textit {$<$man, shirt$>$} being significantly higher than that of \textit {$<$man, elephant$>$}; and \textbf{2)} certain objects are unlikely to appear together in the same scene, and even if they do by a small chance, forming meaningful relationships is still difficult, as exemplified by cases like \textit {$<$airplane, beach$>$}, \textit {$<$giraffe, phone$>$}, \etc. Such prior knowledge of these triplets themselves can clearly reduce the uncertainty in model predictions. Specifically, we can quantify this uncertainty using information entropy $H$: 
\begin{equation}
H(Y_{i j k})=-\sum_{y \in\{1,2, \cdots,  N_{r}\}} P(Y_{i j k}=y) \ln P(Y_{i j k}=y).
\end{equation}
The more uniformly distributed the probability, the higher the information entropy, indicating greater uncertainty about the existence of predicted relationships. However, CCM essentially assumes a uniform distribution of variables and does not incorporate the prior knowledge of the triplets themselves. For example, as shown in Fig. \ref{motivation} (a), ``\textit {man}'' and ``\textit {tree}'' are the two most numerous objects. Under the condition of uniform distribution, the object pair \textit {$<$man, tree$>$} should be the most common. However, in reality, \textit {$<$man, shirt$>$} is the most frequent (see Fig. \ref{motivation} (c)), demonstrating that the $\mathcal{O} \rightarrow \mathcal{P}$ in CCM is insufficient to model the distribution from objects to object pairs.

\begin{figure}
    \footnotesize\centering
    \centerline{\includegraphics[width=1\linewidth]{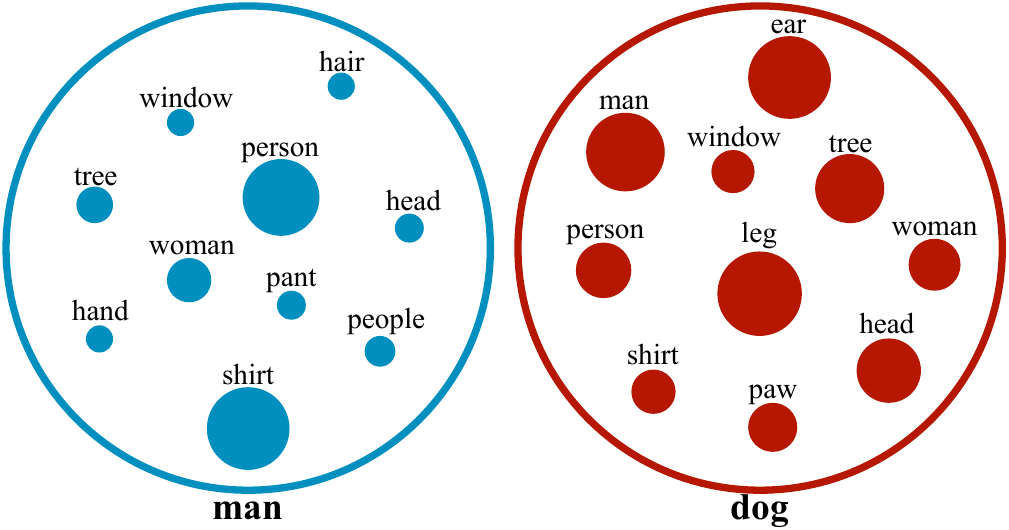}}
        \vspace{-0.2cm}
        \caption{The co-occurrence distributions of two examples (man, dog). For simplicity, only the top 10 co-occurring objects are shown for each example. Solid circles represent the magnitude of co-occurrence, with larger circles indicating a greater likelihood of two objects appearing together in the same scene, and smaller circles suggesting a lower likelihood.}
    \label{contex}
\end{figure}

To optimize $\mathcal{O} \rightarrow \mathcal{P}$, we propose the Mediator-based Causal Chain Model (MCCM), which introduces a mediator variable $\mathcal{C}$ between $\mathcal{O}$ and $\mathcal{P}$. Specifically, $\mathcal{C}=\{\mathcal{C}_{i j}\}_{i=1, j=1}^{N_{o}, N_{o}}$ represents the co-occurrence distributions of any two objects appearing in the same scene. For instance, consider the object pairs \textit{$<$ man, shirt $>$} and \textit{$<$man, tree$>$}. While \textit{man} and \textit{tree} are individually frequent objects in the dataset, their co-occurrence in realistic scenes is far less likely than that of \textit{$<$man, shirt$>$}. In a traditional Causal Chain Model (CCM), the $\mathcal{P}$ distribution relies heavily on the frequencies of individual objects ($\mathcal{O}$), often leading to an overestimation of \textit{$<$man, tree$>$} as a pair. By introducing the mediator variable $\mathcal{C}$, which captures the co-occurrence distributions of object pairs, the model adjusts the $\mathcal{P}$ distribution to reflect the likelihood of real-world scenarios. Specifically, $\mathcal{C}$ assigns a higher weight to \textit{$<$man, shirt$>$}, which frequently co-occurs in realistic contexts, and reduces the weight of \textit{$<$man, tree$>$}, which rarely appears together. This adjustment helps the model focus on meaningful object pairs, thereby reducing bias and improving predictive accuracy. Formally, the co-occurrence between $i$-th and $j$-th object categories can be calculated as:
\begin{equation}
\mathcal{C}_{i j}=\frac{\operatorname{Count}(<r_i, r_j>)}{\operatorname{Count}(r_i)+\operatorname{Count}(r_j)},
\end{equation}
where $\operatorname{Count}(r_j)$ represents the number of object pairs composed of $j$-th object category in observed data, while $\operatorname{Count}(<r_i, r_j>)$ denotes the quantity of object pairs $<r_i, r_j>$. We illustrate the co-occurrence distributions of ``\textit {man}'' and ``\textit {dog}'' in Fig. \ref{contex}. Specifically, from the left subplot in Fig. \ref{contex}, we find $\mathcal{C}_{man,tree} < \mathcal{C}_{man,shirt}$, which explains why ``\textit {man}'' and ``\textit {tree}'' are the most numerous, yet among object pair distributions, \textit {$<$man, shirt$>$} is indeed the most common. The co-occurrence distribution $\mathcal{C}$ can provide prior knowledge of the triplets themselves to reduce the uncertainty in model predictions. We can use conditional mutual information \cite{CCMI,steinke2020reasoning} $I$ to quantify the amount of information that $X_{i j}$ provides for the prediction of $Y_{i j k}$ given $\mathcal{C}_{i j}$:
\begin{equation}
\begin{aligned}
I(Y_{i j k} , X_{i j} \mid \mathcal{C}_{i j})=\sum_{y \in\{1,2, \cdots, N_{r}\}} &\int_{\Omega^2} f_{Y_{i j k}, X_{i j} \mid \mathcal{C}_{i j}}(y, x) \\ &\ln \frac{f_{Y_{i j k}, X_{i j} \mid \mathcal{C}_{i j}}(y, x)}{f_{Y_{i j k} \mid \mathcal{C}_{i j}} (y) f_{X_{i j} \mid \mathcal{C}_{i j}}(x)} d x .
\end{aligned}
\end{equation}
Given the co-occurrence distributions, conditional mutual information can reveal the contribution of $X_{i j}$ to predicting the existence of a relationship; this will be further discussed in Section \ref{sec3.3}. Based on the mediator variable $\mathcal{C}$, the SCM we model is depicted in Fig. \ref{SCM} (b), with its structural equation being: 
\begin{equation}
\begin{aligned}
& \mathcal{O}=P(\mathcal{O}), \\
& \mathcal{C}=\widetilde{F}_\mathcal{O}(\mathcal{O}, P(\mathcal{O})), \\
& \mathcal{P}=\widetilde{F}_1(\mathcal{O}, P(\mathcal{O}))+\widetilde{F}_\mathcal{C}(\mathcal{C}, P(\mathcal{C})), \\
& \mathcal{R}=\widetilde{F}_2(\mathcal{P}, P(\mathcal{P})).
\label{SCM_ours}
\end{aligned}
\end{equation}
$\widetilde{F}_1$, $\widetilde{F}_2$, $\widetilde{F}_\mathcal{O}$ and $\widetilde{F}_\mathcal{C}$ denote structural functions among causal variables. Then, we can get the conditional probability of the outcome variable $\mathcal{R}$:
\begin{equation}
\begin{aligned}
P_{\text {MCCM}}(\mathcal{R} \mid \mathcal{O}, \mathcal{P}, \mathcal{C})= &\sum_{o, p, c} P(\mathcal{R} \mid o, p, c) \cdot P(o \mid \mathcal{O}) \cdot \\ & P(p \mid \mathcal{P}) \cdot P(c \mid \mathcal{C}) .
\end{aligned}
\end{equation}

Compared to the chain structure CCM, our method MCCM employs $\widetilde{F}_\mathcal{O}$ to model the mediator variable $\mathcal{C}$ and utilizes $\widetilde{F}_\mathcal{C}$ to enable $\mathcal{P}$ to avoid object pairs that are unlikely to have actual relationships in reality, thereby guiding the model to focus on meaningful object pairs. In this way, our method yields at least two potential contributions: \textbf{1)} assigning higher weights to object pairs more likely to be found in reality; \textbf{2)} assigning lower weights to those less likely or even impossible to occur. $\widetilde{F}_1$, $\widetilde{F}_2$, $\widetilde{F}_\mathcal{O}$, and $\widetilde{F}_\mathcal{C}$ in Equation (\ref{SCM_ours}) are modeled through our proposed CAModule, as detailed in Section \ref{sec3.3}. 

\begin{figure}
    \footnotesize\centering
    \centerline{\includegraphics[width=0.9\linewidth]{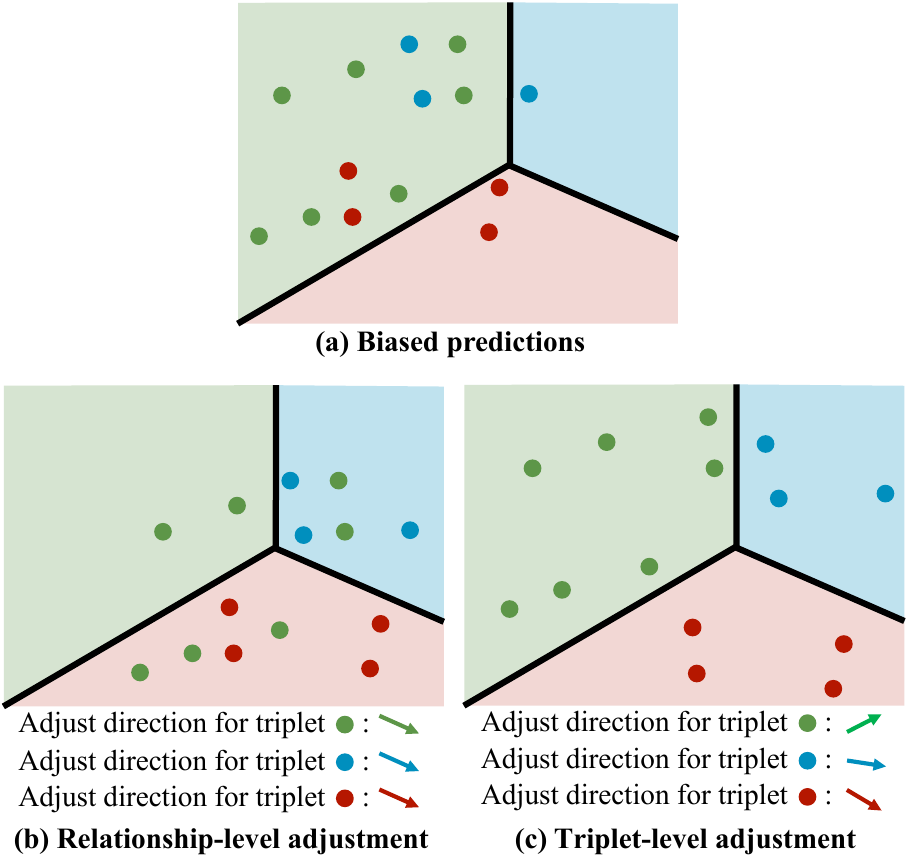}}
        \vspace{-0.2cm}
        \caption{Comparison between relationship-level adjustment and triplet-level adjustment. \protect\tikz \protect\fill[color1] (0,0) circle (0.6ex);, \protect\tikz \protect\fill[color2] (0,0) circle (0.6ex);, and \protect\tikz \protect\fill[color3] (0,0) circle (0.6ex); represent different triplets composed of the same relationship; for illustration, they could be \textit {$<$people, standing on, snow$>$}, \textit {$<$cat, standing on, table$>$}, and \textit {$<$bird, standing on, branch$>$}. (a) shows predictions for the three different triplets \protect\tikz \protect\fill[color1] (0,0) circle (0.6ex);, \protect\tikz \protect\fill[color2] (0,0) circle (0.6ex);, \protect\tikz \protect\fill[color3] (0,0) circle (0.6ex);. (b) illustrates relationship-level adjustment, where since these three triplets belong to the same relationship category, the adjustment direction for all instances is consistent. (c) depicts triplet-level adjustment, where in this finer-grained adjustment, the adjustment direction differs across the different triplets.}
        \label{adjust}
\end{figure}

Within our proposed method based on mediator variables, we can obtain the prediction for $X_{i j}$:
\begin{equation}
f_{Y_{i j k}}(y \mid X_{i j})=P(Y_{i j k}=y \mid \mathcal{O}, \mathcal{C}, \mathcal{P}, \mathcal{R}) .
\end{equation}
Therefore, we can quantify the Average Treatment Effect (ATE) in Equation (\ref{ATE}):
\begin{equation}
E[Y_{i j k} \mid X_{i j}]=\sum_{y \in\{1,2, \cdots,  N_{r}\}} y \cdot f_{Y_{i j k}}(y \mid \mathcal{O}, \mathcal{C}, \mathcal{P}, \mathcal{R}) .
\end{equation}
Note that $f_{Y_{i j k}}(y \mid \mathcal{O}, \mathcal{C}, \mathcal{P}, \mathcal{R})$ represents the outcome after intervening on $\mathcal{P}$ via the mediator variable $\mathcal{C}$, and thus can also be formalized in a counterfactual form: 
\begin{equation}
f_{Y_{i j k}}(y \mid \mathcal{O}, \mathcal{C}, \mathcal{P}, \mathcal{R}) = Y_{i j k}^{d o( \mathcal{P} = p )} - Y_{i j k}^{d o( \mathcal{P} = p_0 )} .
\label{counterfactual-form}
\end{equation}
where $p$ and $p_0$ correspond to two distinct object pair distribution states, with $p$ indicating high-frequency distribution and $p_0$ signifying low frequency. Equation (\ref{counterfactual-form}) shows that our method considers object $o_i$, subject $o_j$, and relationship $r_k$ simultaneously in estimating the ATE, thereby facilitating triplet-level adjustments as outlined in Equation (\ref{triplet-level-adjustment}). However, in traditional relationship-level logit adjustment methods, even diverse triplets with the same relationship correspond to a single adjustment factor. The crucial distinction between relationship-level adjustment and triplet-level adjustment is depicted in Fig. \ref{adjust}. As a result, our approach enables the attainment of the counterfactual outcomes pursued in Equation (\ref{counterfactual}) with one time model inference:
\begin{equation}
\begin{aligned}
Y_{i j k}&^{d o(x)}=  Y_{i j k}^{d o( \mathcal{P} = p )} - Y_{i j k}^{d o( \mathcal{P} = p_0 )}\\
= & \int_o\int_c\int_p\int_r[E(Y_{i j k} \mid X_{i j}, \operatorname{do}(\mathcal{O}=o,\mathcal{C}=c,\mathcal{P}=p,\mathcal{R}=r)) \\
& -E(Y_{i j k} \mid X_{i j}, \operatorname{do}(\mathcal{O}=o,\mathcal{C}=c,\mathcal{P}=p_0,\mathcal{R}=r))] \\
& P(o)P(c)P(p)P(r) do dc dp dr .
\end{aligned}
\end{equation}
The above counterfactuals highlight the intervention in the object pair distribution $\mathcal{P}$. In the following section, we will detail how these counterfactuals are estimated using a lightweight module.

\subsection{Causal Adjustment Module}
\label{sec3.3}
Recent years have seen a surge in interest within the deep learning community in modeling causality using neural networks, with transformers particularly favored for their emphasized capability to capture complex relationships in observed data. In this section, we propose a lightweight causal network based on the transformer architecture \cite{transformer}, the Causal Adjustment Module (CAModule), to model the Mediator-based Causal Chain Model (MCCM) that we introduced in Section \ref{sec3.2}, as shown in Fig. \ref{SCM} (b). CAModule takes as input the causal variables from the MCCM, \ie, $\mathcal{O}$, $\mathcal{C}$, $\mathcal{P}$, $\mathcal{R}$, which are distributions observed in the training data, and its output is a set of adjustment factors to adjust the logits output by the biased SGG models. Specifically, assuming the categories of objects and relationships are $N_{o}$ and $N_{r}$ respectively, the observed distributions include: 

\textbf{1) Object distribution} $\mathcal{O} = \{\mathcal{O}_i\}_{i=1}^{N_{o}} $ $\in \mathbb{R}^{N_{o}}$, where $\mathcal{O}_i$ represents the frequency of $r_i$:
\begin{equation}
\mathcal{O}_i=\frac{\operatorname{Count}(o_i)}{\sum_{i=1}^{N_o} \operatorname{Count}(o_i)} ,
\end{equation}
where $\operatorname{Count}(o_i)$ denotes the count of $o_i$ in the observed data $\mathcal{D}$.

\textbf{2) Co-occurrence distribution} $\mathcal{C} = \{\mathcal{C}_{i j}\}_{i=1, j=1}^{N_{o}, N_{o}}$ $\in \mathbb{R}^{N_{o} \times N_{o} }$, with d denoting the co-occurrence frequency of $r_i$ and $r_j$ within the same scene:
\begin{equation}
\mathcal{C}_{i j}=\frac{\operatorname{Count}(<r_i, r_j>) + \operatorname{Count}(<r_j, r_i>)}{\operatorname{Count}(r_i)+\operatorname{Count}(r_j)} ,
\end{equation}
and $\operatorname{Count}(<r_i, r_j>)$ and $\operatorname{Count}(r_j)$ represent the number of object pair $<r_i, r_j>$ and $r_j$, respectively, in $\mathcal{D}$.

\textbf{3) Object pair distribution} $\mathcal{P} = \{\mathcal{P}_{i j}\}_{i=1, j=1}^{N_{o}, N_{o}}$ $\in \mathbb{R}^{N_{o} \times N_{o} }$, where $\mathcal{P}_{i j}$ signifies the frequency of object pair $<r_i, r_j>$: 
\begin{equation}
\mathcal{P}_{i j}=\frac{\operatorname{Count}(<r_i, r_j>)}{\operatorname{Count}(<,>)} ,
\end{equation}
where $\operatorname{Count}(<,>)$ denotes the total number of object pairs in $\mathcal{D}$.

\textbf{4) Relationship distribution} $\mathcal{R} = \{\mathcal{R}_i\}_{i=1}^{N_{r}}$ $\in \mathbb{R}^{N_{r}}$, with $\mathcal{R}_i$ indicating the frequency of $r_i$:
\begin{equation}
\mathcal{R}_i=\frac{\operatorname{Count}(r_i)}{\sum_{i=1}^{N_r} \operatorname{Count}(r_i)} ,
\label{R-distribution}
\end{equation}
where $\operatorname{Count}(r_i)$ denotes the number of $r_i$ in $\mathcal{D}$.

CAModule first embeds the observed distributions and then captures the causal relationships in the MCCM, including $\mathcal{O} \rightarrow \mathcal{C} \rightarrow \mathcal{R}$ and $\mathcal{O} \rightarrow \mathcal{P} \rightarrow \mathcal{R}$, using the transformer encoder structure $T(\cdot)$ \cite{transformer}, as shown in Algorithm \ref{algorithm1}. Importantly, $\mathcal{O}$, $\mathcal{C}$, $\mathcal{P}$, and $\mathcal{R}$ represent the complete distributions extracted from the observed data $\mathcal{D}$, yet each batch only comprises only a small number of relationship instances. Therefore, for each batch, it is necessary to extract sub-distributions from the complete observed distributions for model training. Assuming batch $\mathcal{B}$ contains $|\mathcal{B}|$ relationship instances, the sub-distributions $\mathcal{O}^{\mathcal{B}}$, $\mathcal{C}^{\mathcal{B}}$, $\mathcal{P}^{\mathcal{B}}$, $\mathcal{R}^{\mathcal{B}}$ are defined as: 
\begin{equation}
\begin{aligned}
& \mathcal{O}^{\mathcal{B}}=[\{\mathcal{O}_i \mid i \in O_{\text {index }}\},\{\mathcal{O}_j \mid j \in S_{\text {index }}\}] \in \mathbb{R}^{|\mathcal{B}| \times 2} , \\
&\mathcal{C}^{\mathcal{B}}=[\{\mathcal{C}_{i j}, \mid i \in O_{\text {index }}, j \in S_{\text {index }}\}\}] \in \mathbb{R}^{|\mathcal{B}|} , \\
&\mathcal{P}^{\mathcal{B}}=[\{\mathcal{P}_{ i j }, \mid i \in O_{\text {index }}, j \in S_{\text {index }}\}\}] \in \mathbb{R}^{|\mathcal{B}|} , \\
& \mathcal{R}^{\mathcal{B}}=\mathcal{R} \in \mathbb{R}^{N_r} ,
\end{aligned}
\end{equation}
where $O_{\text {index }}$ and $S_{\text {index }}$ respectively denote the indexes of the object and subject corresponding to relationship instances in batch $\mathcal{B}$ \renewcommand{\thefootnote}{\fnsymbol{footnote}} \footnote[4]{In the typical two-stage scene graph generation training framework \cite{Neuralmotifs,VCtree}, the object detector $f_o$ is pretrained and remains frozen during the training of the relationship classifier $f_c$; thus, while $f_o$ is used for feature extraction from images, we can also obtain the results of object detection, such as object category information and bounding boxes. Therefore, the indices $O_{\text {index }}$ and $S_{\text {index }}$ for object category are known and can be utilized in the training of classifier $f_c$.}. To capture causal relationships within a unified feature space, sub-distributions $\mathcal{O}^{\mathcal{B}}$, $\mathcal{C}^{\mathcal{B}}$, $\mathcal{P}^{\mathcal{B}}$, and $\mathcal{R}^{\mathcal{B}}$ are initially transformed into feature representations of a consistent dimensionality via linear embedding layers: 
\begin{equation}
\begin{aligned}
& F_{\mathcal{O}}=\operatorname{ReLU}(W_{\mathcal{O}} \cdot \mathcal{O}^{\mathcal{B}}) , \\
& F_{\mathcal{C}}=\operatorname{ReLU}(W_{\mathcal{C}} \cdot \mathcal{C}^{\mathcal{B}}) , \\
& F_{\mathcal{P}}=\operatorname{ReLU}(W_{\mathcal{P}} \cdot \mathcal{P}^{\mathcal{B}}) , \\
& F_{\mathcal{R}}=\operatorname{ReLU}(W_{\mathcal{R}} \cdot \mathcal{R}^{\mathcal{B}}) ,
\end{aligned}
\end{equation}
where $W_{\mathcal{O}}$, $W_{\mathcal{C}}$, $W_{\mathcal{P}}$, and $W_{\mathcal{R}}$ correspond to the weight matrices of the embedding layers for object, co-occurrence, object pair, and relationship distributions, respectively.

\begin{figure}
    \footnotesize\centering
    \centerline{\includegraphics[width=0.9\linewidth]{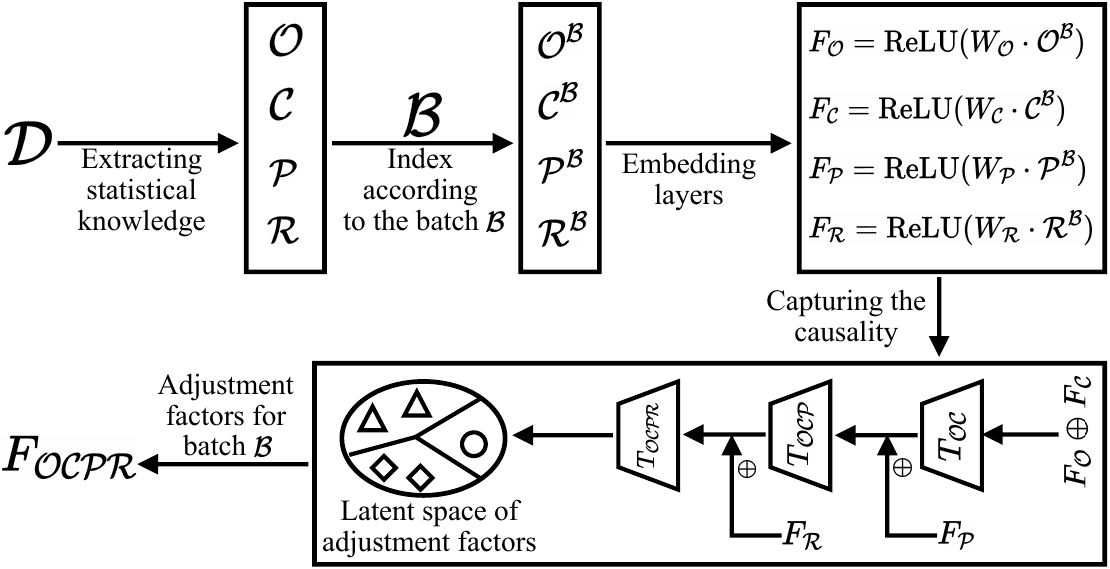}}
        \vspace{-0.2cm}
        \caption{The process of implicitly calculating adjustment factors involves using the skewed distributions $\mathcal{O}$, $\mathcal{C}$, $\mathcal{P}$, and $\mathcal{R}$, extracted from observed data $\mathcal{D}$. It is important to note that only subsets $\mathcal{O}^{\mathcal{B}}$, $\mathcal{C}^{\mathcal{B}}$, $\mathcal{P}^{\mathcal{B}}$, and $\mathcal{R}^{\mathcal{B}}$ of these distributions are involved in each batch.}
        \label{CalculatingAdjustmentFactors}
\end{figure}

\begin{algorithm}[t]
\caption{Transformer Encoder Structure}
\begin{algorithmic}[1]
\State \textbf{Input:} $x$ \qquad \textcolor[RGB]{0,138,115}{// Input tensor of shape $(N, S, E)$, $N$ is the batch size, $S$ is the sequence length, $E$ is the embedding dimension (128 in this case) }
\State \textbf{Parameters:}
\State \quad $W_1, b_1, W_2, b_2$ \qquad \textcolor[RGB]{0,138,115}{// Weight matrix and bias for the linear layers }
\State \quad $\text{LN}_1, \text{LN}_2$ \qquad \textcolor[RGB]{0,138,115}{// Layer normalization }
\Procedure{TransformerEncoderStructure}{$x$}
    \State $x' \gets \Call{MultiHeadAttention}{x, x, x}$
    \State $x'' \gets x + \Call{Dropout}{x'}$
    \State $x'' \gets \Call{LayerNorm}{x'', \text{LN}_1}$
    \State $y \gets \Call{Linear}{x'', W_1, b_1}$
    \State $y \gets \Call{ReLU}{y}$
    \State $y \gets \Call{Dropout}{y}$
    \State $y \gets \Call{Linear}{y, W_2, b_2}$
    \State $y \gets x'' + \Call{Dropout}{y}$
    \State $y \gets \Call{LayerNorm}{y, \text{LN}_2}$
    \State \Return $y$
\EndProcedure
\end{algorithmic}
\label{algorithm1}
\end{algorithm}

Subsequently, we employ a hierarchical feature fusion approach to capture the causal relationships among variables $\mathcal{O}$, $\mathcal{C}$, $\mathcal{P}$, and $\mathcal{R}$. Given $\mathcal{O}$, $\mathcal{C}$, and $\mathcal{P}$, with corresponding components $\mathcal{O} \rightarrow \mathcal{C}$, $\mathcal{O} \rightarrow \mathcal{P}$, and $\mathcal{C} \rightarrow \mathcal{P}$, and acknowledging that $\mathcal{C}$ already incorporates elements of $\mathcal{O}$, we opt to focus solely on $\mathcal{O} \rightarrow \mathcal{C}$ and $\mathcal{C} \rightarrow \mathcal{P}$. Specifically, the embeddings of $\mathcal{O}$ and $\mathcal{C}$ are merged and their causal relationship $\mathcal{O} \rightarrow \mathcal{C}$ is captured through a transformer block: 
\begin{equation}
F_{\mathcal{O C}}=\operatorname{ReLU}(T_{\mathcal{O C}}(F_{\mathcal{O}} \oplus F_{\mathcal{C}})) ,
\end{equation}
\begin{equation}
F_{\mathcal{O C P}}= \operatorname{ReLU}(T_{\mathcal{O C P}}(F_{\mathcal{O C}} \oplus F_{\mathcal{P}}) ) ,
\end{equation}
where $\oplus$ symbolizes the feature concatenation operation. $T_{\mathcal{O C}}$ refers to a transformer block, whose attention mechanism adeptly identifies complex interactions within sequence data, treating the interplay among variables as either sequential or graph-structured data; thus, leads us to believe it can effectively discern the desired causal relationships. Similarly, for $\mathcal{P}$ and $\mathcal{R}$, we merge them and employ a transformer block $T_{\mathcal{C P P R}}$ to capture their causal relationship:
\begin{equation}
F_{\mathcal{O C P R}}=\operatorname{ReLU}(T_{\mathcal{C P P R}}(F_{\mathcal{O C P}} \oplus F_{\mathcal{R}})) ,
\end{equation}
$F_{\mathcal{O C P R}} \in \mathbb{R}^{|\mathcal{B}| \times N_r} $, denoting the adjustment factors corresponding to $|\mathcal{B}|$ relationships in batch $\mathcal{B}$, which, when multiplied by the logits output from the baseline SGG model, yield the final debiasing predictions:
\begin{equation}
P(\mathcal{B}_{i}=k)=\sigma(f_c(f_o(\mathcal{B}_{i})) \times F_{\mathcal{O C P R}} ).
\end{equation}

Fig. \ref{CalculatingAdjustmentFactors} illustrates the process of calculating adjustment factors. In summary, our method initially extracts skewed distributions $\mathcal{O}$, $\mathcal{C}$, $\mathcal{P}$, and $\mathcal{R}$ from observed data $\mathcal{D}$, which are known to induce model bias. We then employ a causal adjustment module that causally models these skewed distributions to obtain adjustment factors for the logits that baseline model output. Notably, our method optimizes the calculation of adjustment factors in an implicit space, in contrast to conventional explicit methods, such as using the inverse of training data frequencies as adjustment factors. Our approach is model-agnostic and can be integrated into SGG model that seeks debiasing predictions. It involves the insertion of a lightweight causal module, \ie, CAModule, for logit adjustment after the baseline model, without any other modifications, including the use of the cross-entropy loss function for supervision, as is typical in baseline models \cite{ImageRetrieval,Neuralmotifs,TDE}:
\begin{equation}
L_{\mathcal{B}}=-\frac{1}{|\mathcal{B}|} \sum_{i=1}^{|\mathcal{B}|} \sum_{k=1}^{N_r} y_{i k} \log (P(\mathcal{B}_i=k)) ,
\end{equation}
where $y_{i k}$ is a one-hot encoded vector representing the true label of the $i$-th data point; $y_{i k} = 1$ if $\mathcal{B}_i$ belongs to the $k$-th category, otherwise $y_{i k} = 0$.

The causal structure modeled by CAModule differs fundamentally from traditional approaches in the modeling from $\mathcal{O}$ to $\mathcal{P}$; specifically, traditional methods is $\mathcal{O} \rightarrow \mathcal{P}$, whereas ours employs $\mathcal{O} \rightarrow \mathcal{C} \rightarrow \mathcal{P}$. Hence, for $\mathcal{O} \rightarrow \mathcal{P}$, the Direct Effect (DE) from $\mathcal{O}$ to $\mathcal{P}$ is: 
\begin{equation}
D E=E[P_i(o) \mid O=o]-E[P_i(o^{\prime}) \mid O=o^{\prime}] ,
\end{equation}
where $P_i$ denotes the potential distribution of $i$-th pairs of objects, with $i$ indexing a specific pair within the scene, potentially any combination of two objects like \textit {$<$cat, table$>$} or \textit {$<$people, chair$>$}. Without any constraints, models naturally tend to represent object pairs based on the frequency of objects. However, this can lead to observations inconsistent with reality, as discussed in Section \ref{sec3.2}, such as ``\textit {man} and \textit {tree} being most frequent, yet \textit {$<$man, shirt$>$} is the most common pair.'' 

In our modeled causal structure MCCM, we introduce the mediator variable $\mathcal{C}$, specifically $\mathcal{O} \rightarrow \mathcal{C} \rightarrow \mathcal{R}$; thus, the Indirect Effect (ME) is:
\begin{equation}
\begin{aligned}
ME_{\mathcal{O}} = & E_{ \mathcal{C} \mid \mathcal{O}=o}[E[\mathcal{P}_i(o, \mathcal{C}) \mid \mathcal{O}=o, \mathcal{C}] \\
& -E[\mathcal{P}_i(o, c^{\prime}) \mid \mathcal{O}=o]] ,
\end{aligned}
\end{equation}
where $E_{\mathcal{C} \mid \mathcal{O}=o}$ represents the expected value of $\mathcal{C}$ given the condition $\mathcal{O}=o$, denoting the average across all possible co-occurrence distribution states. The mediator variable $\mathcal{C}$ emphasizes the incorporation of co-occurrence distribution to constrain object pair distributions. We identify at least three contributions of this to our proposed adjustment module: \textbf{1)} By decomposing the Direct Effect, we can precisely understand the mechanism by which object distribution $\mathcal{O}$ indirectly influence object pair distribution $\mathcal{P}$ through co-occurrence distribution $\mathcal{C}$, leading to a nuanced causal understanding. \textbf{2)} By explicitly modeling the mediating effect, our module can flexibly adjust the weights of different causal pathways to accommodate varying data distributions and task requirements. 3) Direct Effect may struggle to capture rare relationships, however, indirect effects can provide additional information to enhance the prediction of such relationships, which is critical for the SGG task characterized by severe long-tail distributions.

\subsection{Optimize Object Pair Distribution $\mathcal{P}$}
\label{sec3.4}

Our approach achieved the state-of-the-art performance on the critical debiasing metric mR@K, thanks to the CAModule's capability for fine-grained logit adjustments. Surprisingly, we also observed a substantial improvement in the challenging zero-shot relationship recognition metric zR@K (refer to Table~\ref{zRk}), which targets to triplets appearing in the testing set but absent from the training set. This enhancement is believed to stem from our method's decoupling and recombination of relationship triplets, inadvertently generating zero-shot relationships. Specifically, within our proposed MCCM structure, relationships are formed by combining the object pair distribution $\mathcal{P}$ with the relationship distribution $\mathcal{R}$, where $\mathcal{R} = \{\mathcal{R}_i\}_{i=1}^{N_{r}}$ $\in \mathbb{R}^{N_{r}}$ encompasses all relationship categories. As a result, if the training set contains an object pair $<$ $o_m, o_n $ $>$, our method can generate $<$ $o_m, r_i, o_n $ $>$, $i \in[1, N_{r}]$, potentially covering some zero-shot relationships. Notably, while traditional chain causal structure CCM also involve the decoupling and combination of relationships ($\mathcal{O} \rightarrow \mathcal{P}$), these approaches generate $\mathcal{P}$ solely from $\mathcal{O}$ without considering the plausibility of object pair distributions, leading to many implausible triplets. Although this issue persists in our approach, by intervening on $\mathcal{P}$ through the mediator variable $\mathcal{C}$, we effectively downweight these implausible relationships.

Although our method's ability to combine relationships and intervene on the distribution $\mathcal{P}$ enhances recognition of zero-shot relationships, it is crucial to note that such combinations presupposes the presence of object pairs from zero-shot relationships within the training set. In other words, if the object pair of a zero-shot relationship is a zero-shot pair, \ie, absent from the training set but within the testing set, our method's combinatorial potential does not extend to these cases. Addressing this limitation, we further optimize distribution $\mathcal{P}$ in this paper, aiming to encompass as many zero-shot pairs as possible. 

Specifically, we introduce a method for inferring zero-shot pairs from the observed data $\mathcal{D}$, inspired by the prior knowledge that object/subject with similar attributes can form the same relationships. For instance, if \textit {$<$girl, sitting on, chair$>$ } exists in the training set, then \textit {$<$boy, sitting on, chair$>$ } is likely a valid triplet, because both \textit{boy} and \textit{girl}, as beings with similar attributes, are capable of similar actions. Similarly, from \textit {$<$girl, eating, apple$>$ }, we can infer that \textit {$<$girl, eating, pear$>$} is a plausible triplet, since both \textit{apple} and \textit{pear} are fruits with similar attributes. This prior knowledge suggests that for any triplet, the object/subject can be replaced with another object/subject of similar attributes to generate a new, yet reasonable, triplet. Grounded in these insights, we formulate two inference rules:

\textbf{Rule 1}: \textit {If objects $o_m$ and $o_M$ share the same attributes and $<$ $o_m, r_k, o_n $ $>$ is observed in the observed data $\mathcal{D}$, then $<$ $o_M, r_k, o_n $ $>$ is also a reasonable relationship, where $m, M \in [1,N_o]$, $m \neq M$, $k \in [1,N_r]$.}

\textbf{Rule 2}: \textit {If subjects $o_m$ and $o_M$ share the same attributes and $<$ $o_n, r_k, o_m $ $>$ is observed in the observed data $\mathcal{D}$, then $<$ $o_n, r_k, o_M $ $>$ is also a reasonable relationship, where $m, M \in [1,N_o]$, $m \neq M$, $k \in [1,N_r]$.}

\textbf{Rule 1} and \textbf{Rule 2} respectively highlight inference methods when objects and subjects share similar attributes. Consider a simplified structural causal model where relationships are outcomes of objects/subjects and their attributes influenced by causal mechanism $\tilde{f}$; thus, for \textbf{Rule 1} and \textbf{Rule 2}, we have: 
\begin{equation}
r_k=\tilde{f}(o_m, o_M, A_{o_m}, A_{o_M}, U) ,
\end{equation}
where $A_{o_m}$ and $A_{o_M}$ represent the attributes of $o_m$ and $o_M$, respectively. $U$ represents other latent variables, such as the co-occurrence distribution $\mathcal{C}$. \textbf{Rule 1} and \textbf{Rule 2}, which highlight ``If objects $o_m$ and $o_M$ share the same attributes,'' can be interpreted as the causal effect following an intervention on attributes ($d o(A_{o_m})$ or $d o(A_{o_M})$):
\begin{equation}
\begin{split}
P(r_k \mid d o(A_{o_m}=A_{o_M}), o_m, o_M, U) = & \\
P(r_k \mid A_{o_m}, o_m, o_M, U),
\end{split}
\end{equation}
or 
\begin{equation}
\begin{split}
P(r_k \mid d o(A_{o_M}=A_{o_m}), o_m, o_M, U) = & \\
P(r_k \mid A_{o_m}, o_m, o_M, U).
\end{split}
\end{equation}
This suggests that, with all else being equal, altering the attributes of $o_m$ to match those of $o_M$ does not alter the distribution of relationship $r_k$, assuming the attributes of both objects are equivalent in their influence on $r_k$. 

To quantify the attribute similarity between $o_m$ and $o_M$, we formally define the similarity between them as $S_{o_m, o_M}$, viewed as an effort to estimate conditional independence with given observed data $\mathcal{D}$. Thus, the observed attributes can be utilized to approximate the unobserved causal effect: 
\begin{equation}
S_{o_m, o_M} \approx P(r_k \mid d o(A_{o_m}), d o(A_{o_M}), o_m, o_M, U) .
\end{equation}
To compute $S_{o_m, o_M}$, we initially extract features of object instances using language model GloVe \cite{Glove} \renewcommand{\thefootnote}{\fnsymbol{footnote}} \footnote[4]{GloVe \cite{Glove} is an unsupervised learning algorithm for obtaining vector representations for words. The prevalent SGG frameworks universally employ pretrained language models for embedding object classifications, with GloVe being a frequent choice. Therefore, our use of GloVe to calculate the similarity of object attributes does not introduce additional external knowledge, ensuring a fair comparison.}. Assuming two object instances belong to categories $o_m$ and $O_M$, and their features extracted by GloVe model are denoted as $\mathbf{v}_{o_m}$ and $\mathbf{v}_{o_M}$. Although various methods exist for calculating distances between features, such as Euclidean, Manhattan, and cosine similarity, for $\mathbf{v}_{o_m}$ and $\mathbf{v}_{o_M}$, two considerations are essential: \textbf{1)} the absolute distance between the features in a high-dimensional space; \textbf{2)} their directional similarity, since objects in scene graph task are directional, for example, \textit {$<$girl, chair$>$ } and \textit {$<$chair, girl$>$ } represent entirely different object pairs. Given these considerations, we propose using both Euclidean distance and cosine similarity to compute the distance between $\mathbf{v}_{o_m}$ and $\mathbf{v}_{o_M}$:
\begin{equation}
d(\mathbf{v}_{o_m}, \mathbf{v}_{o_M})=\|\mathbf{v}_{o_m}-\mathbf{v}_{o_M}\| ,
\end{equation}
\begin{equation}
\cos (\theta_{\mathbf{v}_{o_m}, \mathbf{v}_{o_M}})=\frac{\mathbf{v}_{o_m} \cdot \mathbf{v}_{o_M}}{\|\mathbf{v}_{o_m}\|\|\mathbf{v}_{o_M}\|} .
\end{equation}
Euclidean distance $d(\mathbf{v}_{o_m}, \mathbf{v}_{o_M})$ captures differences in magnitude, while cosine similarity $\cos (\theta_{\mathbf{v}_{o_m}, \mathbf{v}_{o_M}})$ reflects semantic differences. We then weight these two distances:
\begin{equation}
S_{o_m, o_M}=\alpha \cdot \cos (\theta_{\mathbf{v}_{o_m}, \mathbf{v}_{o_M}})+(1-\alpha) \cdot(1-\frac{d(\mathbf{v}_{o_m}, \mathbf{v}_{o_M})}{1+d(\mathbf{v}_{o_m}, \mathbf{v}_{o_M})}),
\label{distance-calculation}
\end{equation}
where $\alpha$ is a hyperparameter used for weighting. By substituting $o_m$ and $o_M$ in Equation (\ref{distance-calculation}), we can get the attribute similarity between any two objects. Finally, we set a threshold $\beta$ to assess whether two objects are similar.

Experimental results (see Fig. \ref{alpha}) indicate that the model performs optimally when $\alpha = 0.7$, highlighting the critical role of directionality in assessing object attribute similarity. However, further increasing $\alpha$ leads to a decline in model performance, signifying the importance of absolute distance in high-dimensional space. Regarding threshold $\beta$ (see Fig. \ref{beta}), a lower threshold can recall nearly all zero-shot object pairs but may result in many implausible pairs. Conversely, a higher threshold yields mostly plausible object pairs but risks missing numerous zero-shot pairs. It is evident that the optimization of the object pair distribution $\mathcal{P}$ is contingent on computing the similarity of object attributes, which is dependent on the specific language model employed. Therefore, it might be beneficial to reassess the optimal values for $\alpha$ and $\beta$ when employing different language models. For further analysis of $\alpha$ and $\beta$, refer to Section \ref{Ablations}.   

\subsection{Compared to Current Debiasing Method}

In Section \ref{Related-works-SGG}, we discuss established debiasing approaches including resampling, reweighting, adjustment, and hybrid methods. Resampling methods are prone to overfitting (oversample) or underfitting (undersample), while reweighting methods may lead to poor and unstable performance due to sensitivity in model parameter updates, often reducing the effectiveness of certain categories, especially prominent ones. Among these, Adjustment methods are most similar to ours. However, our method presents key distinctions:

\textbf{1)} Our method comprehensively addresses skewed distributions that bias the model, including object distribution $\mathcal{O}$, co-occurrence distribution $\mathcal{C}$, object pair distribution $\mathcal{P}$, and relationship distribution $\mathcal{R}$, whereas other methods typically focus only on $\mathcal{R}$. By addressing these interconnected distributions, our approach mitigates bias at a more foundational level, enhancing the robustness of scene graph generation (SGG) models.

\textbf{2)} Our adjustment parameters are at the triplet level, crucial for SGG tasks, since, unlike relationship-level methods, triplet-level adjustments account for the long-tail distribution even within identical relationships. This allows our approach to more effectively handle the inherent complexity of SGG tasks, leading to improved performance on tail relationships without sacrificing head category accuracy.

\textbf{3)} Using causal modeling techniques, our method optimizes adjustment parameters in an implicit space, producing parameters applicable to any triplet, thus enabling generalization to zero-shot relationships. This contrasts with common methods that explicitly compute parameters, such as the reciprocal of category distributions.

\begin{table*}[htbp]
  \centering
  \vspace{-0.3cm}
  \caption{Quantitative results on mR@K metric. $\text{AVG}_{\text{mR}}^{\Delta}$ denotes the mean of mR@20, mR@50, and mR@100, while $\text{AVG}_{\text{mR}}^{\Diamond}$ is the mean of mR@50 and mR@100. The best results are marked in blue.}
    \resizebox{19.5cm}{!}{\begin{tabular}{c|l|ccccc|ccccc|ccccc}
    \toprule
          &   & \multicolumn{5}{c|}{PredCls}  & \multicolumn{5}{c|}{SGCls}    & \multicolumn{5}{c}{SGDet} \\
          &   & \multicolumn{1}{c}{mR@20} & \multicolumn{1}{c}{mR@50} & \multicolumn{1}{c}{mR@100} & \multicolumn{1}{c}{$\text{AVG}_{\text{mR}}^{\Delta}$} & \multicolumn{1}{c|}{$\text{AVG}_{\text{mR}}^{\Diamond}$} & \multicolumn{1}{c}{mR@20} & \multicolumn{1}{c}{mR@50} & \multicolumn{1}{c}{mR@100} & \multicolumn{1}{c}{$\text{AVG}_{\text{mR}}^{\Delta}$} & \multicolumn{1}{c|}{$\text{AVG}_{\text{mR}}^{\Diamond}$}  &\multicolumn{1}{c}{mR@20} & \multicolumn{1}{c}{mR@50} & \multicolumn{1}{c}{mR@100} & \multicolumn{1}{c}{$\text{AVG}_{\text{mR}}^{\Delta}$} & \multicolumn{1}{c}{$\text{AVG}_{\text{mR}}^{\Diamond}$} \\
    \midrule
    \midrule
    \multirow{31}{*}{\rotatebox{90}{VG150 \cite{VG150}}} 
     & MotifsNet (backbone) \cite{Neuralmotifs}  & $12.2$  & $15.5$  & $16.8$  & $14.8$  & $16.2$  & $7.2$   & $9.0$     & $9.5$   & $8.6$ & $9.3$  & $5.2$   & $7.2$   & $8.5$   & $7.0$ & $7.9$ \\
    & \quad $\text {HML \cite{HML}}$  \scriptsize \textit {(ECCV’22)} & $30.1$  & $36.3$  & $38.7$  & $35.0$ & $37.5$ & $17.1$  & $20.8$  & $22.1$  & $20.0$ & $21.5$  & $10.8$  & $14.6$  & $17.3$  & $14.2$ & $16.0$  \\
    & \quad $\text {FGPL \cite{FGPL}}$  \scriptsize \textit {(CVPR’22)}  & $24.3$  & $33.0$  & $37.5$  & $31.6$ & $35.3$ & $17.1$  & $21.3$  & $22.5$  & $20.3$ & $21.9$ & $11.1$  & $15.4$  & $18.2$  & $14.9$  & $16.8$ \\
    & \quad $\text {TransRwt \cite{TransRwt}}$  \scriptsize \textit {(ECCV’22)}  & $-$     & $35.8$  & $39.1$  & $-$ & $37.5$ & $-$     & $\textcolor{blue}{21.5}$  & $22.8$  & $-$ & $22.2$ & $-$     & $15.8$  & $18.0$  & $-$ & $16.9$  \\
    & \quad $\text {PPDL \cite{PPDL}}$  \scriptsize \textit {(CVPR’22)}  & $-$     & $32.2$  & $33.3$  & $-$  & $32.8$  & $-$     & $17.5$  & $18.2$  & $-$  & $17.9$  & $-$     & $11.4$  & $13.5$  & $-$ & $12.5$  \\
    & \quad $\text {GCL \cite{GCL}}$  \scriptsize \textit {(CVPR’22)}  & $30.5$  & $36.1$  & $38.2$  & $34.9$ & $37.2$ & $18.0$    & $20.8$  & $21.8$  & $20.2$  & $21.3$  & $\textcolor{blue}{12.9}$  & $\textcolor{blue}{16.8}$  & $\textcolor{blue}{19.3}$  & $\textcolor{blue}{16.3}$ & $\textcolor{blue}{18.1}$ \\
     & \quad $\text {NICEST \cite{NICEST}}$  \scriptsize \textit {(CVPR’22)}  & $-$     & $30.0$    & $32.1$  & $-$ & $31.1$  & $-$     & $16.4$  & $17.5$  & $-$  & $17.0$ & $-$     & $10.4$  & $12.7$  & $-$ & $11.6$  \\
    & \quad $\text {RTPB \cite{RTPB}}$  \scriptsize \textit {(AAAI’22)}  & $28.8$  & $35.3$  & $37.7$  & $33.9$ & $36.5$  & $16.3$  & $19.4$  & $22.6$  & $19.4$  & $21.0$ & $9.7$   & $13.1$  & $15.5$  & $12.8$ & $14.3$ \\
    & \quad $\text {TBE \cite{TBE}}$  \scriptsize \textit {(CVPR’23)}   &$ -  $&$  24.7 $&$ 30.7  $&$ -  $ & $ 27.7$  &$ -  $&$ 14.5  $&$ 17.4  $&$ -  $  & $ 16.0 $  &$ - $&$  9.4  $&$  11.7  $&$ -$ & $ 10.6 $ \\ 
    & \quad $\text {CFA \cite{CFA}}$  \scriptsize \textit {(ICCV’23)}   &$ -  $&$ 35.7  $&$ 38.2  $&$ -  $ & $37.0$  &$ -  $&$ 17.0  $&$ 18.4  $&$ -  $  & $17.7$  &$ - $&$ 13.2  $&$ 15.5  $&$ -$ & $14.4$ \\
    & \quad $\text {DKBL \cite{DKBL}}$  \scriptsize \textit {(MM’23)}   &$ -  $&$ 29.7  $&$ 32.2  $&$ -  $ & $30.1$  &$ -  $&$ 18.2  $&$ 19.4  $&$ -  $  & $18.8$  &$ - $&$ 12.6  $&$ 15.1  $&$ -$ & $13.9$ \\
    & \quad $\text {ST-SGG \cite{ST-SGG}}$  \scriptsize \textit {(ICLR’24)}   &$ -  $&$ 32.5  $&$ 35.1  $&$ -  $ & $33.8$  &$ -  $&$ 18.0  $&$ 19.3  $&$ -  $  & $18.7$  &$ - $&$ 12.9  $&$ 15.8  $&$ -$ & $14.4$ \\
    & \quad \textbf{CAModule (ours)}   &  $\textcolor{blue}{32.5} $    &   $ \textcolor{blue}{36.7}$    &   $\textcolor{blue}{39.3} $    &  $\textcolor{blue}{36.2} $     & $ \textcolor{blue}{38.0}$      &   $\textcolor{blue}{18.8} $    &   $21.1 $    &   $\textcolor{blue}{24.7} $    &   $\textcolor{blue}{21.5} $    &   $\textcolor{blue}{22.9} $    &  $11.7$     & $16.3 $  &   $ 18.2$    &   $15.4 $    & $17.3 $ \\
    \cline{2-17} \noalign{\vskip 1mm}
    & VCTree (backbone) \cite{VCtree} &$12.4$ 	&$15.4$ 	&$16.6$ 	&$14.8$ &$16.0$ &$6.3$ 	&$7.5$ 	&$8.0$ 	&$7.3$  &$7.8$ &$4.9$ 	&$6.6$ 	&$7.7$ 	&$6.4$ &$7.2$  \\
    & \quad $\text {HML \cite{HML}}$  \scriptsize \textit {(ECCV’22)}  &$31.0 	$&$36.9 	$&$39.2 	$&$35.7 	$ &$30.1$ &$20.5 	$&$25.0 	$&$26.8 	$&$24.1 	$ &$25.9$ &$10.1 	$&$13.7 	$&$16.3 	$&$13.4$ &$15.0$  \\
    & \quad $\text {FGPL \cite{FGPL}}$  \scriptsize \textit {(CVPR’22)}  &$30.8 	$&$37.5 	$&$40.2 	$&$36.2 	$ &$38.9$ &$\textcolor{blue}{21.9} 	$&$\textcolor{blue}{26.2} 	$&$\textcolor{blue}{27.6} 	$&$\textcolor{blue}{25.2} 	$ &$\textcolor{blue}{26.9}$ &$11.9 	$&$16.2 	$&$19.1 	$&$15.7$ &$17.7$   \\
    & \quad $\text {TransRwt \cite{TransRwt}}$  \scriptsize \textit {(ECCV’22)} &$- 	$&$37.0 	$&$39.7 	$&$- 	$ &$38.4$ &$- 	$&$19.9 	$&$21.8 	$&$- 	$ &$20.9$ &$- 	$&$12.0 	$&$14.9 	$&$-$  &$13.5$ \\
    & \quad $\text {PPDL \cite{PPDL}}$  \scriptsize \textit {(CVPR’22)} &$- 	$&$33.3 	$&$33.8 	$&$- 	$ &$33.6$ &$- 	$&$21.8 	$&$22.4 	$&$- 	$ &$22.1$ &$- 	$&$11.3 	$&$13.3 	$&$-$ &$12.3$  \\
    & \quad $\text {GCL \cite{GCL}}$  \scriptsize \textit {(CVPR’22)} &$31.4 	$&$37.1 	$&$39.1 	$&$35.9 	$ &$37.5$ &$19.5 	$&$22.5 	$&$23.5 	$&$21.8 	$ &$23.0$ &$11.9 	$&$15.2 	$&$17.5 	$&$14.9$ &$16.4$   \\
     & \quad $\text {NICEST \cite{NICEST}}$  \scriptsize \textit {(CVPR’22)} &$- 	$&$30.9 	$&$33.1 	$&$- 	$ &$32.0$ &$- 	$&$20.0 	$&$21.2 	$&$- 	$ &$20.6$ &$- 	$&$10.1 	$&$12.1 	$&$-$ &$11.1$   \\
    & \quad $\text {RTPB \cite{RTPB}}$  \scriptsize \textit {(AAAI’22)}  &$27.3 	$&$33.4 	$&$35.6 	$&$32.1 	$ &$34.5$ &$20.6 	$&$24.5 	$&$25.8 	$&$23.6 	$ &$25.2$ &$9.6 	$&$12.8 	$&$15.1 	$&$12.5$ &$14.0$  \\
    & \quad $\text {TBE \cite{TBE}}$  \scriptsize \textit {(CVPR’23)}   &$ -  $&$ 28.1  $&$ 30.7  $&$ -  $ & $ 29.4$  &$ -  $&$  17.3 $&$  19.4 $&$ -  $  & $ 18.4 $  &$ - $&$   10.4 $&$  11.9  $&$ -$ & $ 11.2 $ \\  
    & \quad $\text {CFA \cite{CFA}}$  \scriptsize \textit {(ICCV’23)}   &$ -  $&$ 34.5  $&$ 37.2  $&$ -  $ & $35.6$  &$ -  $&$ 19.1  $&$ 20.8  $&$ -  $  & $20.0$  &$ - $&$ 13.1  $&$ 15.5  $&$ -$ & $14.3$ \\
    & \quad $\text {DKBL \cite{DKBL}}$  \scriptsize \textit {(MM’23)}   &$ -  $&$ 28.7  $&$ 31.3  $&$ -  $ & $30.0$  &$ -  $&$ 21.2  $&$ 22.6  $&$ -  $  & $21.9$  &$ - $&$ 11.8  $&$ 14.2  $&$ -$ & $13.0$ \\
    & \quad $\text {ST-SGG \cite{ST-SGG}}$  \scriptsize \textit {(ICLR’24)}   &$ -  $&$ 32.7  $&$  35.6  $&$ -  $ & $34.2$  &$ -  $&$ 21.0  $&$  22.4  $&$ -  $  & $21.7$  &$ - $&$ 12.6  $&$  15.1  $&$ -$ & $13.9$ \\ 
    & \quad \textbf{CAModule (ours)}   &  $ \textcolor{blue}{35.2}$    &   $\textcolor{blue}{38.4} $    &   $\textcolor{blue}{40.5} $    &  $ \textcolor{blue}{38.0}$     & $\textcolor{blue}{39.5} $      &   $ 21.7$    &   $23.2 $    &   $ 25.9$    &   $ 23.6$    &   $ 24.6$    &  $\textcolor{blue}{12.3} $     & $\textcolor{blue}{16.4} $  &   $\textcolor{blue}{19.6} $    &   $\textcolor{blue}{16.1} $    & $ \textcolor{blue}{18.5}$ \\
    \cline{2-17} \noalign{\vskip 1mm}
    
    & Transformer (backbone) \cite{transformer} &$ 12.4  $&$ 16.0  $&$ 17.5  $&$ 15.3  $ &$16.8$ &$ 7.7   $&$ 9.6     $&$ 10.2   $&$ 9.2   $ &$9.9$ &$ 5.3   $&$ 7.3   $&$ 8.8   $&$ 7.1$ &$8.1$  \\
    & \quad $\text {HML \cite{HML}}$  \scriptsize \textit {(ECCV’22)} &$ 30.1  $&$ 36.3  $&$ 38.7  $&$35.0  $ &$37.5$ &$ \textcolor{blue}{17.1}  $&$ 20.8  $&$ 22.1  $&$ 20.0  $ &$21.5$ &$ 10.8  $&$ 14.6  $&$ 17.3  $&$ 14.2$ &$16.0$  \\
    & \quad $\text {FGPL \cite{FGPL}}$  \scriptsize \textit {(CVPR’22)}  &$ 24.3  $&$ 33.0  $&$ 37.5  $&$ 31.6  $ &$35.3$ &$ \textcolor{blue}{17.1}  $&$ 21.3  $&$ 22.5  $&$ 20.3  $ &$21.9$ &$ 11.1  $&$ 15.4  $&$ 18.2  $&$ 14.9$ &$16.8$  \\
    & \quad $\text {TransRwt \cite{TransRwt}}$  \scriptsize \textit {(ECCV’22)}  &$ -     $&$ 35.8  $&$ 39.1  $&$ -  $ &$37.5$ &$ -     $&$ \textcolor{blue}{21.5}  $&$ 22.8  $&$ -  $ &$22.2$ &$ -     $&$ 15.8  $&$ 18.0  $&$ -$ &$16.9$  \\
    & \quad $\text {CFA \cite{CFA}}$  \scriptsize \textit {(ICCV’23)}   &$ -  $&$ 30.1  $&$ 33.7  $&$ -  $ & $31.9$  &$ -  $&$ 15.7  $&$ 17.2  $&$ -  $  & $16.5$  &$ - $&$ 12.3  $&$ 14.6  $&$ -$ & $13.5$ \\
    & \quad $\text {DKBL \cite{DKBL}}$  \scriptsize \textit {(MM’23)}   &$ -  $&$ 30.0  $&$ 33.3  $&$ -  $ & $31.7$  &$ -  $&$ 17.7  $&$ 19.4  $&$ -  $  & $18.6$  &$ - $&$ 13.2  $&$ 15.8  $&$ -$ & $14.5$ \\    
    & \quad \textbf{CAModule (ours)}   &  $ \textcolor{blue}{34.1}$    &   $\textcolor{blue}{37.6} $    &   $\textcolor{blue}{39.4} $    &  $\textcolor{blue}{37.0} $     & $\textcolor{blue}{38.5} $      &   $ 16.7$    &   $ 21.3$    &   $\textcolor{blue}{23.9} $    &   $ \textcolor{blue}{20.6}$    &   $\textcolor{blue}{22.6} $    &  $\textcolor{blue}{12.2} $     & $ \textcolor{blue}{16.6}$  &   $ \textcolor{blue}{19.3}$    &   $\textcolor{blue}{16.0} $    & $ \textcolor{blue}{18.0}$ \\

    \midrule
    \midrule
    \multirow{13}{*}{\rotatebox{90}{GQA \cite{GQAdataset}}} & MotifsNet (backbone) \cite{Neuralmotifs}  &$11.7$& $15.4$ & $16.3$  & $14.5$  & $15.9$  & $6.8$   & $8.6$     & $9.4$   & $8.3$ & $9.0$  & $4.4$   & $6.8$   & $8.6$   & $6.6$ & $7.7$ \\
    & \quad GCL \cite{GCL} \scriptsize \textit {(CVPR’22)} &-  & $36.7$  & $38.1$  & -  & $37.4$  & -   & $17.3$     & $18.1$   & - & $17.7$  & -   & $16.8$   & $\textcolor{blue}{18.8}$   & - & $\textcolor{blue}{17.8}$ \\
    & \quad CFA \cite{CFA} \scriptsize \textit {(ICCV’23)}  &-  & $31.7$ & $33.8$  & -  & $32.8$  & -   & $14.2$     & $15.2$   & - & $14.7$  & -   & $11.6$   & $13.2$   & - & $12.4$ \\
    & \quad EICR \cite{EICR} \scriptsize \textit {(ICCV’23)}  &-  & $ 36.3$ & $38.0$  & -  & $37.2$  & -   & $17.2$     & $18.2$   & - & $17.7$  & -   & $ 16.0$   & $18.0$   & - & $17.0$ \\
    & \quad \textbf{CAModule (ours)}   &$\textcolor{blue}{31.7} $ &$\textcolor{blue}{37.2} $ &$ \textcolor{blue}{38.7}$&$ \textcolor{blue}{35.9}$ & $ \textcolor{blue}{38.0}$  & $ \textcolor{blue}{12.9}$   & $\textcolor{blue}{19.1}$     & $\textcolor{blue}{20.6}$   & $\textcolor{blue}{17.5}$ & $\textcolor{blue}{19.9}$  & $\textcolor{blue}{11.9}$   & $\textcolor{blue}{16.9}$   & $18.4$   & $\textcolor{blue}{15.7}$ & $17.7$ \\
    \cline{2-17} \noalign{\vskip 1mm}
    & VCTree (backbone) \cite{VCtree} &$11.9$  & $14.3$ & $16.2$  & $14.1$  & $15.3$  & $6.8$   & $7.3$     & $7.8$   & $7.3$ & $7.6$  & $3.8$   & $6.1$   & $7.4$   & $5.7$ & $6.8$ \\
    & \quad GCL \cite{GCL} \scriptsize \textit {(CVPR’22)}  &-  & $35.4$ & $36.7$  & -  & $36.1$  & -   & $17.3$     & $18.0$   & - & $17.7$  & -   & $15.6$   & $17.8$   & - & $16.7$ \\
    & \quad CFA \cite{CFA} \scriptsize \textit {(ICCV’23)} &-  & $33.4$ & $35.1$  & -  & $34.3$  & -   & $14.1$     & $15.0$   & - & $14.6$  & -   & $10.8$   & $12.6$   & - & $11.7$ \\
    & \quad EICR \cite{EICR} \scriptsize \textit {(ICCV’23)}  &-  & $35.9$ & $37.4$  & -  & $36.7$  & -   & $17.8$     & $18.6$   & - & $18.2$  & -   & $ 14.7$   & $16.3$   & - & $15.5$ \\
    & \quad \textbf{CAModule (ours)}   &$\textcolor{blue}{33.6}$ & $\textcolor{blue}{37.7}$ & $\textcolor{blue}{39.8}$  & $\textcolor{blue}{37.0}$  & $\textcolor{blue}{38.8}$  & $\textcolor{blue}{14.0}$   & $\textcolor{blue}{19.6}$     & $\textcolor{blue}{21.7}$   & $\textcolor{blue}{18.4}$ & $\textcolor{blue}{20.7}$  & $\textcolor{blue}{12.0}$   & $\textcolor{blue}{17.3}$   & $\textcolor{blue}{19.7}$   & $\textcolor{blue}{16.3}$ & $\textcolor{blue}{18.5}$ \\
    \cline{2-17} \noalign{\vskip 1mm}
    & Transformer (backbone) \cite{transformer} &$12.3$  & $15.2$ & $17.7$  & $15.1$  & $16.5$  & $7.3$   & $8.4$     & $9.6$   & $8.4$ & $9.0$  & $5.1$   & $6.6$   & $7.9$   & $6.5$ & $7.3$ \\
    & \quad CFA \cite{CFA} \scriptsize \textit {(ICCV’23)}  &-  & $27.8$ & $29.4$  & -  & $28.6$  & -   & $16.2$     & $16.9$   & - & $16.6$  & -   & $13.4$   & $15.3$   & - & $14.4$ \\
    & \quad \textbf{CAModule (ours)}   &$\textcolor{blue}{33.9}$  & $\textcolor{blue}{38.3}$ & $\textcolor{blue}{40.2}$  & $\textcolor{blue}{37.5}$  & $\textcolor{blue}{39.3}$  & $\textcolor{blue}{13.7}$   & $\textcolor{blue}{18.8}$     & $\textcolor{blue}{20.8}$   & $\textcolor{blue}{17.8}$ & $\textcolor{blue}{19.8}$  & $\textcolor{blue}{12.3}$  & $\textcolor{blue}{17.5}$   & $\textcolor{blue}{19.6}$   & $\textcolor{blue}{16.5}$ & $\textcolor{blue}{18.6}$ \\
    \midrule
    \midrule
    \end{tabular}}%
  \label{mRK}%
  \vspace{-0.2cm}
\end{table*}%

\section{Experiments}
\label{Experiments}

\subsection{Experiment Setup}
\label{Experiment-Setup}

\textit {Implementations.} 
We evaluate our CAModule on the three datasets, VG150 \cite{VG150}, GQA \cite{GQAdataset}, and Open Images V6 (OI V6) \cite{openimages}, as described below.

\textbf{VG150} \cite{VG150} is a subset of the VG dataset \cite{VG} comprising 150 object categories and 50 relationship categories. We follow the split in \cite{TDE}, consisting of 62k images for training, 5k for validation, and 26k for testing.

\textbf{GQA} \cite{GQAdataset} is a large-scale visual question answering dataset with real images from the Visual Genome dataset and balanced question-answer pairs. Following the configuration in \cite{GCL}, we selecte the top-200 object classes and top-100 relationship categories, adhering to its split of 70\% for training (including a 5k validation set) and 30\% for testing.

\textbf{OI V6} \cite{openimages} is a large dataset designed for tasks like image classification, object detection, and visual relationship detection. Following the settings in \cite{BGNN}, it includes 601 object categories and 30 relationship categories, with 126k images for training, 2k for validation, and 5k for testing.

All experiments are conducted using the TDE's open-source repository \cite{TDE}, adhering to most of its settings: \textbf{1)} The object detector $f_o$, a Faster R-CNN \cite{fasterrcnn} with a ResNeXt-101-FPN backbone \cite{resnet}, achieved a 28.14 mAP on the VG test set and remains frozen during training of the classifier $f_c$. \textbf{2)} We utilize three popular SGG frameworks, MotifsNet \cite{Neuralmotifs}, VCTree \cite{VCtree}, and Transformer \cite{transformer}, with batch sizes and initial learning rates of 12 and 0.01 for MotifsNet and VCTree, and 16 and 0.001 for Transformer, respectively. \textbf{3)} CAModule primarily involves two hyperparameters, $\alpha$ and $\beta$, controlling the weights for Euclidean distance and cosine similarity, and the similarity threshold between two objects, set at 0.7 and 0.4, respectively. A detailed discussion of these hyperparameters is provided in Section \ref{Ablations}.

\textit {Backbones and baselines.} 
The experiments are conducted across three widely-used SGG backbones: MotifsNet \cite{Neuralmotifs}, VCTree \cite{VCtree}, and Transformer \cite{transformer}. The baseline approaches primarily include debiasing techniques developed in the past two years, encompassing resampling (\eg, TransRwt \cite{TransRwt}), reweighting (\eg, PPDL \cite{PPDL}), adjustment (\eg, RTPB \cite{RTPB}), and hybrid (\eg, NICEST \cite{NICEST}) methods.

\textit {Evaluation modes.}
Following \cite{Neuralmotifs,VCtree,TDE}, we employ three standard evaluation modes: \textbf{1)} Predicate classification (PredCls), where the SGG model is tasked with predicting relationships between object pairs given their category and location information. \textbf{2)} Scene Graph Classification (SGCls), which requires predicting both the category of objects and their relationships based on given location information. \textbf{3)} Scene Graph Detection (SGDet), where the SGG model must predict the category, location of objects, and their relationships without any prior object-related information. Generally, the difficulty level increases progressively across these three modes, leading to a corresponding stepwise decrease in model performance.

\textit {Evaluation metrics.}
For \textbf{VG150} \cite{VG150} and \textbf{GQA} \cite{GQAdataset}, we employ three evaluation metrics: \textbf{1)} Recall rate (R@K). R@K has been a prevalent metric in the early SGG domain, where for each scene, the top $K$ predictions with the highest confidence are selected, and the proportion of correct predictions is calculated. Typically, $K$ is set to 20, 50, and 100, \ie, R@20, R@50, and R@100, respectively. \textbf{2)} mean Recall rate (mR@K). mR@K has become increasingly popular in recent years, especially for debiasing SGG task, where it has emerged as a principal metric. Unlike R@K, mR@K calculates the recall rate for each relationship category and then averages these rates. \textbf{3)} Mean of R@K and mR@K (MR@K). R@K is a category-agnostic metric, thus allowing models to achieve high scores by excelling in only a few head categories. Overemphasis on mR@K, however, can compromise performance in these head categories, which are often more prevalent in real-world contexts, potentially limiting the model's utility in practical applications. In response,  MR@K computes the average of R@K and mR@K, with a high MR@K indicating a model has effectively balanced performance across both head and tail categories, thereby ensuring a more holistic evaluation.

For \textbf{OI V6} \cite{openimages}, we present results using the metric of $\text{R@50}$ and $\text{mR@50}$, along with the weighted mean AP of relationships ($\text{wmAP}_{rel}$) and weighted mean AP of phrase ($\text{wmAP}_{phr}$). Consistent with the established evaluation protocols \cite{BGNN,PCL,Runet}, we also report the scorewtd metric, calculated as: $\text{score}_{wtd} = 0.2 \times \text{R@50} + 0.4 \times \text{wmAP}_{rel} + 0.4 \times \text{wmAP}_{phr}$.

\begin{table*}[htbp]
  \centering
  \caption{Quantitative results on R@K metric. $\text{AVG}_{\text{R}}^{\Delta}$ and $\text{AVG}_{\text{R}}^{\Diamond}$ share similar meanings to those described in Table~\ref{mRK}.}
    \resizebox{19.5cm}{!}{\begin{tabular}{l|ccccc|ccccc|ccccc}
    \toprule
          & \multicolumn{5}{c|}{PredCls}  & \multicolumn{5}{c|}{SGCls}    & \multicolumn{5}{c}{SGDet} \\
          & \multicolumn{1}{c}{R@20} & \multicolumn{1}{c}{R@50} & \multicolumn{1}{c}{R@100} & \multicolumn{1}{c}{$\text{AVG}_{\text{R}}^{\Delta}$} & \multicolumn{1}{c|}{$\text{AVG}_{\text{R}}^{\Diamond}$} & \multicolumn{1}{c}{R@20} & \multicolumn{1}{c}{R@50} & \multicolumn{1}{c}{R@100} & \multicolumn{1}{c}{$\text{AVG}_{\text{R}}^{\Delta}$} & \multicolumn{1}{c|}{$\text{AVG}_{\text{R}}^{\Diamond}$}  &\multicolumn{1}{c}{R@20} & \multicolumn{1}{c}{R@50} & \multicolumn{1}{c}{R@100} & \multicolumn{1}{c}{$\text{AVG}_{\text{R}}^{\Delta}$} & \multicolumn{1}{c}{$\text{AVG}_{\text{R}}^{\Diamond}$} \\
    \midrule
    \midrule
    MotifsNet (backbone) \cite{Neuralmotifs} & $59.5$&$66.0	$&$67.9	$&$64.5 	$ &$67.0$ &$35.8	$&$39.1	$&$39.9	$&$38.3 	$ &$39.5$ &$25.1	$&$32.1	$&$36.9	$&$31.4$ &$34.5$ \\
    \quad $\text {TransRwt \cite{TransRwt}}$  \scriptsize \textit {(ECCV’22)} &$-$&$48.6	$&$50.5	$&$- 	$ &$49.6$ &$-	$&$29.4	$&$30.2	$&$- 	$ &$29.8$ &$-	$&$23.5	$&$27.2 	$&$-$ &$25.4$  \\
    \quad $\text {PPDL \cite{PPDL}}$  \scriptsize \textit {(CVPR’22)} &$- $&$47.2 $&$47.6 $&$- $ &$47.4$ &$- $&$28.4 $&$29.3 $&$- $ &$28.9$ &$- $&$21.2 $&$23.9 $&$-$ &$22.6$ \\
    \quad $\text {TBE \cite{TBE}}$  \scriptsize \textit {(CVPR’23)}   &$ -  $&$ 51.5  $&$ 55.1  $&$ -  $ & $ 53.3$  &$ -  $&$  32.2 $&$ 33.8  $&$ -  $  & $ 33.0 $  &$ - $&$  23.9  $&$  27.1  $&$ -$ & $25.5 $ \\  
        \quad $\text {CFA \cite{CFA}}$  \scriptsize \textit {(ICCV’23)}   &$ -  $&$ 54.1  $&$ 56.6  $&$ -  $ & $55.4$  &$ -  $&$ 34.9 $&$ 36.1  $&$ -  $  & $35.5$  &$ - $&$ 24.7  $&$ 31.8  $&$ -$ & $29.6$ \\
    \quad $\text {DKBL \cite{DKBL}}$  \scriptsize \textit {(MM’23)}   &$ -  $&$ 57.2  $&$ 58.8  $&$ -  $ & $58.0$  &$ -  $&$ 32.7  $&$ 33.4  $&$ -  $  & $33.1$  &$ - $&$ 27.0  $&$ 30.7  $&$ -$ & $28.9$ \\  
    \quad $\text {ST-SGG \cite{ST-SGG}}$  \scriptsize \textit {(ICLR’24)}   &$ -  $&$ 50.5 $&$ 52.8 $&$ -  $ & $51.7$  &$ -  $&$ 31.2  $&$ 32.1  $&$ -  $  & $31.7$  &$ - $&$ 25.7  $&$ 29.8 $&$ -$ & $27.8$ \\
    \quad \textbf{CAModule (ours)}   &  $\textcolor{blue}{55.9} $    &   $ \textcolor{blue}{59.8}$    &   $\textcolor{blue}{63.4} $    &  $\textcolor{blue}{59.7} $     & $\textcolor{blue}{61.6} $      &   $ \textcolor{blue}{30.1}$    &   $ \textcolor{blue}{36.8}$    &   $\textcolor{blue}{38.2} $    &   $ \textcolor{blue}{35.0}$    &   $\textcolor{blue}{37.5} $    &  $ \textcolor{blue}{22.7}$     & $ \textcolor{blue}{29.1}$  &   $\textcolor{blue}{32.7} $    &   $\textcolor{blue}{28.2} $    & $ \textcolor{blue}{30.9}$ \\
    \midrule
    \midrule
    VCTree (backbone) \cite{VCtree} &$59.8	$&$66.2	$&$68.1	$&$64.7 	$ &$67.2$ &$37.0	$&$40.5	$&$41.4	$&$39.6 	$ &$41.0$ &$24.7	$&$31.5	$&$36.2	$&$30.8$ &$33.9$   \\
    \quad $\text {TransRwt \cite{TransRwt}}$  \scriptsize \textit {(ECCV’22)} &$-	$&$48.0	$&$49.9	$&$- 	$ &$49.0$ &$-	$&$30.0	$&$30.9	$&$- 	$ &$30.5$ &$-	$&$23.6	$&$27.8	$&$-$ &$25.7$  \\
    \quad $\text {PPDL \cite{PPDL}}$  \scriptsize \textit {(CVPR’22)} &$-$&$47.6$&$48	$&$- $ &$47.8$ &$-$&$32.1	$&$33$&$- $ &$32.6$ &$-	$&$20.1	$&$22.9	$&$-$ &$21.5$ \\
    \quad $\text {TBE \cite{TBE}}$  \scriptsize \textit {(CVPR’23)}   &$ -  $&$  59.5 $&$  61.0 $&$ -  $ & $ 60.3$  &$ -  $&$ 40.7  $&$ 41.6  $&$ -  $  & $ 41.2$  &$ - $&$  27.7  $&$  30.1  $&$ -$ & $ 28.9$ \\  
     \quad $\text {CFA \cite{CFA}}$  \scriptsize \textit {(ICCV’23)}   &$ -  $&$ 54.7  $&$ 57.2  $&$ -  $ & $ 56.1 $  &$ -  $&$ \textcolor{blue}{42.4} $&$ \textcolor{blue}{43.5}  $&$ -  $  & $\textcolor{blue}{43.0}$  &$ - $&$ 27.1  $&$ 31.2  $&$ -$ & $29.2$ \\ 
    \quad $\text {DKBL \cite{DKBL}}$  \scriptsize \textit {(MM’23)}   &$ -  $&$ 60.1  $&$ 61.8  $&$ -  $ & $61.0$  &$ -  $&$ 38.8  $&$ 39.7  $&$ -  $  & $39.3$  &$ - $&$ 26.9  $&$ 30.7  $&$ -$ & $28.8$ \\
    \quad $\text {ST-SGG \cite{ST-SGG}}$  \scriptsize \textit {(ICLR’24)}   &$ -  $&$ 52.5 $&$ 54.3 $&$ -  $ & $53.4$  &$ -  $&$ 36.3  $&$  37.3  $&$ -  $  & $36.8$  &$ - $&$ 20.7  $&$ 24.9  $&$ -$ & $22.8$ \\
    \quad \textbf{CAModule (ours)}   &  $ \textcolor{blue}{56.3}$    &   $\textcolor{blue}{60.6} $    &   $\textcolor{blue}{63.8} $    &  $\textcolor{blue}{60.2} $     & $\textcolor{blue}{62.2} $      &   $\textcolor{blue}{30.8} $    &   $38.6 $    &   $40.2 $    &   $\textcolor{blue}{36.5} $    &   $39.4 $    &  $ \textcolor{blue}{23.1}$     & $\textcolor{blue}{29.8} $  &   $ \textcolor{blue}{33.4}$    &   $\textcolor{blue}{28.8} $    & $\textcolor{blue}{31.6}$ \\
    \midrule
    \midrule
    Transformer (backbone) \cite{transformer} &$58.5$&$65.0	$&$66.7	$&$63.4 	$ &$66.0$ &$35.6	$&$38.9	$&$39.8	$&$38.1 	$ &$39.4$ &$24.0	$&$30.3	$&$33.3	$&$29.2$ &$31.8$  \\
    \quad $\text {TransRwt \cite{TransRwt}}$  \scriptsize \textit {(ECCV’22)} &$-$&$49.0	$&$50.8	$&$- $ &$49.9$ &$-	$&$29.6	$&$30.5	$&$- 	$ &$30.1$ &$-	$&$23.1	$&$27.1	$&$-$  &$25.1$ \\
         \quad $\text {CFA \cite{CFA}}$  \scriptsize \textit {(ICCV’23)}   &$ -  $&$ 59.2  $&$ 61.5  $&$ -  $ & $60.4$  &$ -  $&$ 36.3 $&$ 37.3  $&$ -  $  & $36.8$  &$ - $&$ 27.7  $&$ 32.1  $&$ -$ & $29.9$ \\
    \quad $\text {DKBL \cite{DKBL}}$  \scriptsize \textit {(MM’23)}   &$ -  $&$ 57.5  $&$ 59.1  $&$ -  $ & $58.3$  &$ -  $&$ 33.2  $&$ 34.0  $&$ -  $  & $33.6$  &$ - $&$ 28.0  $&$ 31.7  $&$ -$ & $29.9$ \\
    \quad \textbf{CAModule (ours)}   &  $\textcolor{blue}{55.5} $    &   $ \textcolor{blue}{59.4}$    &   $\textcolor{blue}{61.7} $    &  $ \textcolor{blue}{58.9}$     & $\textcolor{blue}{60.6} $      &   $ \textcolor{blue}{29.3}$    &   $\textcolor{blue}{36.4} $    &   $\textcolor{blue}{37.9} $    &   $ \textcolor{blue}{34.5}$    &   $\textcolor{blue}{37.2} $    &  $\textcolor{blue}{22.6} $     & $\textcolor{blue}{28.6} $  &   $\textcolor{blue}{32.8} $    &   $\textcolor{blue}{28.0} $    & $\textcolor{blue}{30.7} $ \\
    \bottomrule
    \end{tabular}}%
  \label{RK}%
\end{table*}%

\begin{table*}[htbp]
  \centering
  \caption{Quantitative results on MR@K metric. ${\text{MR@K}}^{\Delta}$ and ${\text{MR@K}}^{\Diamond}$ share similar meanings to those described in Table~\ref{mRK}.}
    \resizebox{19.5cm}{!}{\begin{tabular}{l|cccccc|cccccc|cccccc}
    \toprule
          & \multicolumn{6}{c|}{PredCls}  & \multicolumn{6}{c|}{SGCls}    & \multicolumn{6}{c}{SGDet} \\
          & \multicolumn{1}{c}{$\text{AVG}_{\text{mR}}^{\Delta}$} & \multicolumn{1}{c}{$\text{AVG}_{\text{R}}^{\Diamond}$} & \multicolumn{1}{c}{$\text{AVG}_{\text{mR}}^{\Delta}$} & \multicolumn{1}{c}{$\text{AVG}_{\text{R}}^{\Diamond}$} & \multicolumn{1}{c}{${\text{MR@K}}^{\Delta}$} & \multicolumn{1}{c|}{${\text{MR@K}}^{\Diamond}$} & \multicolumn{1}{c}{$\text{AVG}_{\text{mR}}^{\Delta}$} & \multicolumn{1}{c}{$\text{AVG}_{\text{R}}^{\Diamond}$} & \multicolumn{1}{c}{$\text{AVG}_{\text{mR}}^{\Delta}$} & \multicolumn{1}{c}{$\text{AVG}_{\text{R}}^{\Diamond}$} & \multicolumn{1}{c}{${\text{MR@K}}^{\Delta}$} & \multicolumn{1}{c|}{${\text{MR@K}}^{\Diamond}$} &\multicolumn{1}{c}{$\text{AVG}_{\text{mR}}^{\Delta}$} & \multicolumn{1}{c}{$\text{AVG}_{\text{R}}^{\Diamond}$} & \multicolumn{1}{c}{$\text{AVG}_{\text{mR}}^{\Delta}$} & \multicolumn{1}{c}{$\text{AVG}_{\text{R}}^{\Diamond}$} & \multicolumn{1}{c}{${\text{MR@K}}^{\Delta}$} & \multicolumn{1}{c}{${\text{MR@K}}^{\Diamond}$} \\
    \midrule
    \midrule
    MotifsNet (backbone) \cite{Neuralmotifs} &$14.8$ &$16.2$ &$64.5$ &$67.0$ &$39.7$ &$41.6$ &$8.6$ &$9.3$ &$38.3$ &$39.5$ &$23.5$ &$24.4$ &$7.0$ &$7.9$ &$31.4$ &$34.5$ &$19.2$ &$21.2$  \\
    \quad $\text {TransRwt \cite{TransRwt}}$  \scriptsize \textit {(ECCV’22)} &$-$ &$37.5$ &$-$ &$49.6$ &$-$ &$43.6$  &$-$ &$22.2$ &$-$ &$29.8$ &$-$ &$26.0$ &$-$ &$16.9$ &$-$ &$25.4$ &$-$ &$21.2$    \\
    \quad $\text {PPDL \cite{PPDL}}$  \scriptsize \textit {(CVPR’22)} &$-$ &$32.8$ &$-$ &$47.4$ &$-$ &$40.1$ &$-$ &$17.9$ &$-$ &$28.9$ &$-$ &$23.4$ &$-$ &$12.5$ &$-$ &$22.6$ &$-$ &$17.6$ \\
    \quad $\text {TBE \cite{TBE}}$  \scriptsize \textit {(CVPR’23)}   &   $-$    &    $27.7$   &    $-$   &   $53.5$    &   $-$    &    $40.5$   &   $-$    &   $16.0$    &  $-$     &   $33.0$    &   $-$    & $24.5$  &   $-$    &    $10.6$   &    $-$   &   $25.5$    &   $-$    &    $18.1$  \\
    \quad $\text {CFA \cite{CFA}}$  \scriptsize \textit {(ICCV’23)}   &$ -  $&$ 37.0  $&$ -  $&$ 55.4  $ & $-$  &$ 46.2  $&$ - $&$ 17.7  $&$ -  $  & $35.5$  &$ - $&$ 26.6  $&$ -  $&$ 14.4$ & $-$ & $29.6$ & $-$ & $22.0$\\
    \quad $\text {DKBL \cite{DKBL}}$  \scriptsize \textit {(MM’23)}   &    $-$ &   $30.1$    &    $-$   &    $58.0$   &   $-$    &   $44.1$    &    $-$   &   $18.8$    &   $-$    &  $33.1$     &   $-$    &   $26.0$    & $-$  &   $13.9$    &    $-$   &    $28.9$   &   $-$    &   $21.4$   \\
    \quad $\text {ST-SGG \cite{ST-SGG}}$  \scriptsize \textit {(ICLR’24)}   &$ -  $&$ 33.8 $&$ - $&$ 51.7  $ & $-$  &$ 42.8  $&$ -  $&$  18.7  $&$ -  $  & $31.7$  &$ - $&$ 25.2  $&$ -  $&$ 14.4$ & $-$ & $27.8$ & $-$  & $21.1$ \\
    \quad \textbf{CAModule (ours)}   &   $\textcolor{blue}{36.2}$    &    $\textcolor{blue}{38.0}$   &    $\textcolor{blue}{59.7}$   &   $\textcolor{blue}{61.6}$    &   $\textcolor{blue}{48.0}$    &    $\textcolor{blue}{49.8}$   &   $\textcolor{blue}{21.5}$    &   $\textcolor{blue}{22.9}$    &  $\textcolor{blue}{35.0}$     &   $\textcolor{blue}{37.5}$    &   $\textcolor{blue}{28.3}$    & $\textcolor{blue}{30.2}$  &   $\textcolor{blue}{15.4}$    &    $\textcolor{blue}{17.3}$   &    $\textcolor{blue}{28.2}$   &   $\textcolor{blue}{30.9}$    &   $\textcolor{blue}{21.8}$    &    $\textcolor{blue}{24.1}$  \\
    \midrule
    \midrule
    VCTree (backbone) \cite{VCtree} &$14.8$ &$16.0$ &$64.7$ &$67.2$ &$39.8$ &$41.6$ &$7.3$ &$7.8$ &$39.6$ &$41.0$ &$23.5$ &$24.4$ &$6.4$ &$7.2$ &$30.8$ &$33.9$ &$18.6$ &$20.6$    \\
    \quad $\text {TransRwt \cite{TransRwt}}$  \scriptsize \textit {(ECCV’22)} &$-$ &$38.4$ &$-$ &$49.0$ &$-$ &$43.7$ &$-$ &$20.9$ &$-$ &$30.5$ &$-$ &$25.7$ &$-$ &$13.5$ &$-$ &$25.7$ &$-$ &$19.6$  \\
    \quad $\text {PPDL \cite{PPDL}}$  \scriptsize \textit {(CVPR’22)} &$-$ &$33.6$ &$-$ &$47.8$ &$-$ &$40.7$ &$-$ &$22.1$ &$-$ &$32.6$ &$-$ &$27.4$ &$-$ &$12.3$ &$-$ &$21.5$ &$-$ &$16.9$ \\
    \quad $\text {TBE \cite{TBE}}$  \scriptsize \textit {(CVPR’23)}   &   $-$    &    $29.4$   &    $-$   &   $60.3$    &   $-$    &    $44.9$   &   $-$    &   $18.4$    &  $-$     &   $41.2$    &   $-$    & $29.8$  &   $-$    &    $11.2$   &    $-$   &   $28.9$    &   $-$    &    $20.1$  \\
             \quad $\text {CFA \cite{CFA}}$  \scriptsize \textit {(ICCV’23)}   &$ -  $&$ 35.6  $&$ -  $&$ 56.1  $ & $-$  &$ 45.9  $&$ - $&$ 20.0  $&$ -  $  & $43.0$  &$ - $&$ 31.5  $&$ -  $&$ 14.3$ & $-$ & $29.2$ & $-$ & $21.8$\\
    \quad $\text {DKBL \cite{DKBL}}$  \scriptsize \textit {(MM’23)}   &   $-$    &    $30.0$   &    $-$   &   $61.0$    &   $-$    &    $45.5$   &   $-$    &   $21.9$    &  $-$     &   $39.3$    &   $-$    & $30.6$  &   $-$    &    $13.0$   &    $-$   &   $28.8$    &   $-$    &    $20.9$  \\
    \quad $\text {ST-SGG \cite{ST-SGG}}$  \scriptsize \textit {(ICLR’24)}   &$ -  $&$ 34.2 $&$ - $&$ 53.4  $ & $-$  &$ 43.8  $&$ -  $&$  21.7  $&$ -  $  & $36.8$  &$ - $&$ 29.3 $&$ -  $&$ 13.9$ & $-$ & $22.8$ & $-$  & $18.4$ \\
    \quad \textbf{CAModule (ours)}   &   $\textcolor{blue}{38.0}$    &    $\textcolor{blue}{39.5}$   &    $\textcolor{blue}{60.2}$   &   $\textcolor{blue}{62.2}$    &   $\textcolor{blue}{49.1}$    &    $\textcolor{blue}{50.9}$   &   $\textcolor{blue}{23.6}$    &   $\textcolor{blue}{24.6}$    &  $\textcolor{blue}{36.5}$     &   $\textcolor{blue}{39.4}$    &   $\textcolor{blue}{30.1}$    & $\textcolor{blue}{32.0}$  &   $\textcolor{blue}{16.1}$    &    $\textcolor{blue}{18.5}$   &    $\textcolor{blue}{28.8}$   &   $\textcolor{blue}{31.6}$    &   $\textcolor{blue}{22.5}$    &    $\textcolor{blue}{25.1}$  \\
    \midrule
    \midrule
    Transformer (backbone) \cite{transformer} &$15.3$ &$16.8$ &$63.4$ &$66.0$ &$39.4$ &$41.4$ &$9.2$ &$9.9$ &$38.1$ &$39.4$ &$23.7$ &$24.7$ &$7.1$ &$8.1$ &$29.2$ &$31.8$ &$18.2$ &$20.0$  \\
    \quad $\text {TransRwt \cite{TransRwt}}$  \scriptsize \textit {(ECCV’22)} &$-$ &$37.5$ &$-$ &$49.9$ &$-$ &$43.7$ &$-$ &$22.2$ &$-$ &$30.1$ &$-$ &$26.2$ &$-$ &$16.9$ &$-$ &$25.1$ &$-$ &$21.0$ \\
    \quad $\text {CFA \cite{CFA}}$  \scriptsize \textit {(ICCV’23)}   &$ -  $&$ 31.9  $&$ -  $&$ 60.4  $ & $-$  &$ 46.2  $&$ - $&$ 16.5  $&$ -  $  & $36.8$  &$ - $&$ 26.7  $&$ -  $&$ 13.5$ & $-$ & $29.9$ & $-$ & $21.7$\\
    \quad $\text {DKBL \cite{DKBL}}$  \scriptsize \textit {(MM’23)}   &   $-$    &    $31.7$   &    $-$   &   $58.3$    &   $-$    &    $45.0$   &   $-$    &   $18.6$    &  $-$     &   $33.6$    &   $-$    & $26.1$  &   $-$    &    $14.5$   &    $-$   &   $29.9$    &   $-$    &    $22.2$  \\
    \quad \textbf{CAModule (ours)}   &   $\textcolor{blue}{37.0}$    &    $\textcolor{blue}{38.5}$   &    $\textcolor{blue}{58.9}$   &   $\textcolor{blue}{60.6}$    &   $\textcolor{blue}{48.0}$    &    $\textcolor{blue}{49.6}$   &   $\textcolor{blue}{20.6}$    &   $\textcolor{blue}{22.6}$    &  $\textcolor{blue}{34.5}$     &   $\textcolor{blue}{37.2}$    &   $\textcolor{blue}{27.6}$    & $\textcolor{blue}{29.9}$  &   $\textcolor{blue}{16.0}$    &    $\textcolor{blue}{18.0}$   &    $\textcolor{blue}{28.0}$   &   $\textcolor{blue}{30.7}$    &   $\textcolor{blue}{22.0}$    &    $\textcolor{blue}{24.4}$  \\
    \bottomrule
    \end{tabular}}%
  \label{MRK}%
\end{table*}%

\begin{table}[htbp]
  \centering
  \vspace{-0.3cm}
  \caption{Quantitative results of RcSGG and baseline methods on OI V6 dataset. The SGG backbone used here is Transformer \cite{transformer}. }
    \begin{tabular}{lccccc}
    \toprule
    \multicolumn{1}{c}{\multirow{2}[4]{*}{Models}} & \multicolumn{1}{r}{\multirow{2}[4]{*}{mR@50}} & \multicolumn{1}{r}{\multirow{2}[4]{*}{R@50}} & \multicolumn{2}{c}{wmAP} & \multicolumn{1}{c}{\multirow{2}[4]{*}{$\text{score}_{wtd}$}} \\
\cmidrule{4-5}          &       &       & \multicolumn{1}{l}{$rel$} & \multicolumn{1}{l}{$phr$} &  \\
    \midrule
    BGNN \cite{BGNN}  \scriptsize \textit {(CVPR’21)}   &   40.5   & 75.0     & 33.5     & 34.2     & 42.1 \\
    PCL \cite{PCL} \scriptsize \textit {(TIP’22)}     &   41.6   & 74.8     & 34.7     & 35.0     & 42.8 \\
    RU-Net \cite{Runet} \scriptsize \textit {(CVPR’22)} &   -   & 76.9     & 35.4     & 34.9     & 43.5 \\
    PE-NET \cite{PENET} \scriptsize \textit {(CVPR’23)} &   -   & 76.5     & 36.6     & 37.4     & 44.9 \\
    FGPL-A \cite{FGPL-A} \scriptsize \textit {(TPAMI’23)} &   -  & 73.4     & 35.9     & 36.8     & 43.8 \\
    HetSGG \cite{HetSGG} \scriptsize \textit {(AAAI’23)} &   43.2  & 74.8     & 33.5     & 34.5     & 42.2 \\
    SQUAT \cite{SQUAT} \scriptsize \textit {(CVPR’23)} &   -   & 75.8     & 34.9     & 35.9    & 43.5 \\
    IWSL \cite{IWSL} \scriptsize \textit {(TPAMI’24)} &   42.2   & 74.7     & 33.1     & 34.3    & 41.9 \\
    \midrule
    \textbf{CAModule (ours)} &   $\textcolor{blue}{43.7}$   & $\textcolor{blue}{77.4}$     & $\textcolor{blue}{40.3}$     & $\textcolor{blue}{43.0}$    & $\textcolor{blue}{48.8}$ \\
    
    \bottomrule
    \end{tabular}%
  \label{OpenImagesV6}%
  \vspace{-0.3cm}
\end{table}%

\subsection{Comparison with State-of-the-art}
\label{results}
In this section, we evaluate our method from both quantitative and qualitative analyses.

\textit {Quantitative analyses.}
Quantitative results are reported in Table~\ref{mRK},~\ref{RK},~\ref{MRK}, and ~\ref{OpenImagesV6}, covering the metrics mR@K, R@K, MR@K, wmAP, and $\text{score}_{wtd}$, respectively. These results show that:

\textbf{1)} Table~\ref{mRK} showcases that our method achieves state-of-the-art performance on the crucial debiasing metric mR@K for scene graph generation task on VG150 and GQA datasets, with similar results observed for OI V6 in Table~\ref{OpenImagesV6}. For instance, in the PredCls mode with VCTree \cite{VCtree} as the backbone, our method outperforms resampling methods (\eg, TransRwt \cite{TransRwt}), reweighting methods (\eg, PPDL \cite{PPDL}), adjustment methods (\eg, RTPB \cite{RTPB}), and hybrid methods (\eg, NICEST \cite{NICEST}) on mR@100, with improvements of 2.7\%, 8.6\%, 6.8\%, and 9.3\%, respectively. For GQA, with MotifsNet as the backbone, the improvement ranges from 1.9\% (TransRwt \cite{TransRwt}) to 7.3\% (TBE \cite{TBE}) under the SGCls mode withe same metric. On OI V6, our method outperforms others by 0.5\% (HetSGG \cite{HetSGG}) to 3.2\% (BGNN \cite{BGNN}) on the mR@50 metric. Given that some baseline methods have missing numbers, we also calculate metrics $\text{AVG}_{\text{mR}}^{\Delta}$ and $\text{AVG}_{\text{mR}}^{\Diamond}$ in Table~\ref{mRK} for a more comprehensive comparison, under which our approach continues to demonstrate superior performance.

\begin{table*}[htbp]
  \centering
  \caption{Quantitative results on zR@K metric. $\text{CAModule}^{-\mathcal{P}}$ denotes model performance without optimizing the object pair distribution $\mathcal{P}$. $\text{AVG}_{\text{zR}}^{\Delta}$ and $\text{AVG}_{\text{zR}}^{\Diamond}$ share similar meanings to those described in Table~\ref{mRK}.}
    \resizebox{19.5cm}{!}{\begin{tabular}{l|ccccc|ccccc|ccccc}
    \toprule
          & \multicolumn{5}{c|}{PredCls}  & \multicolumn{5}{c|}{SGCls}    & \multicolumn{5}{c}{SGDet} \\
          & \multicolumn{1}{c}{zR@20} & \multicolumn{1}{c}{zR@50} & \multicolumn{1}{c}{zR@100} & \multicolumn{1}{c}{$\text{AVG}_{\text{zR}}^{\Delta}$} & \multicolumn{1}{c|}{$\text{AVG}_{\text{zR}}^{\Diamond}$} & \multicolumn{1}{c}{zR@20} & \multicolumn{1}{c}{zR@50} & \multicolumn{1}{c}{zR@100} & \multicolumn{1}{c}{$\text{AVG}_{\text{zR}}^{\Delta}$} & \multicolumn{1}{c|}{$\text{AVG}_{\text{zR}}^{\Diamond}$}  &\multicolumn{1}{c}{zR@20} & \multicolumn{1}{c}{zR@50} & \multicolumn{1}{c}{zR@100} & \multicolumn{1}{c}{$\text{AVG}_{\text{zR}}^{\Delta}$} & \multicolumn{1}{c}{$\text{AVG}_{\text{zR}}^{\Diamond}$} \\
    \midrule
    \midrule
    MotifsNet (backbone) \cite{Neuralmotifs}  & $4.2$  & $7.3$  & $9.4$  & $7.0$  & $5.8$  & $0.2$   & $0.7$     & $1.7$   & $0.9$ & $1.2$  & $0.0$   & $0.0$   & $0.3$   & $0.1$ & $0.2$ \\ 
    \quad \textbf{$\text{CAModule}^{-\mathcal{P}}$}   &  $ 7.0$    &   $10.8$    &   $13.5$    &  $10.4$     & $12.2$      &   $2.1$    &   $3.4$    &   $4.9$    &   $3.5$    &   $4.2$    &  $0.3$     & $1.1$  &   $2.5$    &   $1.3$    & $1.8$ \\
    \quad \textbf{CAModule}   &  $9.5 $    &   $16.7 $    &   $20.3 $    &  $15.5 $     & $18.5 $      &   $2.1 $    &   $5.1 $    &   $7.6 $    &   $ 4.9$    &   $6.4 $    &  $1.4 $     & $2.1 $  &   $3.7 $    &   $2.4 $    & $2.9 $ \\
    \midrule
    \midrule
    VCTree (backbone) \cite{VCtree} &$4.3$ 	&$7.7$ 	&$10.3$ 	&$7.4$ &$9.0$ &$0.4$ 	&$0.7$ 	&$2.1$ 	&$1.1$  &$1.4$ &$0.0$ 	&$0.0$ 	&$0.5$ 	&$0.2$ &$0.3$  \\
    \quad \textbf{$\text{CAModule}^{-\mathcal{P}}$}   &  $6.6$    &   $11.3$    &   $13.8$    &  $10.6$     & $12.6$      &   $2.0$    &   $3.4$    &   $4.7$    &   $3.4$    &   $4.1$    &  $0.3$     & $0.9$  &   $2.3$    &   $1.2$    & $1.6$ \\
    \quad \textbf{CAModule}   &  $ 8.1$    &   $14.5 $    &   $18.4 $    &  $13.7$     & $ 16.5$      &   $1.3 $    &   $5.1 $    &   $ 7.2$    &   $ 4.5$    &   $6.2 $    &  $ 1.1$     & $1.6 $  &   $ 3.2$    &   $ 2.0$    & $2.4 $ \\
    \midrule
    \midrule
    Transformer (backbone) \cite{transformer} &$ 4.1  $&$ 7.9  $&$ 11.3  $&$ 7.8  $ &$9.6$ &$ 0.3   $&$ 0.7     $&$ 2.3   $&$ 1.1   $ &$1.5$ &$ 0.0   $&$ 0.0   $&$ 0.4   $&$ 0.1$ &$0.2$  \\ 
    \quad \textbf{$\text{CAModule}^{-\mathcal{P}}$}   &  $6.4$    &   $9.6$    &   $13.1$    &  $9.7$     & $11.4$      &   $1.9$    &   $3.2$    &   $4.2$    &   $3.1$    &   $3.7$    &  $0.2$     & $0.7$  &   $2.1$    &   $1.0$    & $1.4$ \\
    \quad \textbf{CAModule}   &  $ 8.5$    &   $15.4 $    &   $19.7 $    &  $14.5 $     & $17.6 $      &   $1.5 $    &   $ 4.8$    &   $8.4$    &   $ 4.9$    &   $6.6 $    &  $ 1.2$     & $ 1.7$  &   $4.6 $    &   $2.5 $    & $ 3.2$ \\
    \bottomrule
    \end{tabular}}%
  \label{zRk}%
\end{table*}%

\begin{figure*}[htbp]
    \footnotesize\centering
    \centerline{\includegraphics[width=0.95\linewidth]{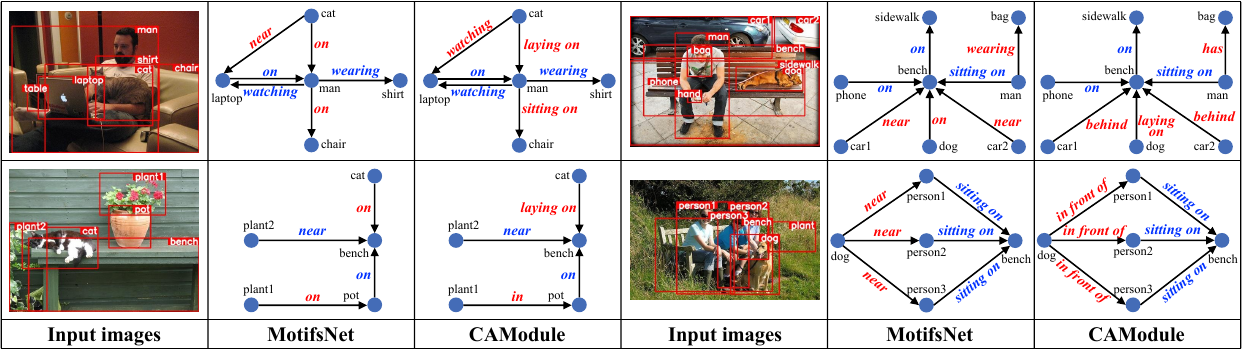}}
        \vspace{-0.2cm}
        \caption{Qualitative results of our method compared to the baseline approach (MotifsNet). Same relationship predictions are highlighted in blue, while differing predictions are emphasized in red.}
        \label{Qualitative}
\end{figure*}

\textbf{2)} Table~\ref{RK} illustrates that our approach maintains its state-of-the-art performance on the VG150 dataset across the R@K metric compared to debiasing baselines. For instance, in the SGCls mode with MotifsNet \cite{Neuralmotifs} as the backbone, our approach surpasses debiasing methods like TransRwt \cite{TransRwt}, PPDL \cite{PPDL}, and TBE \cite{TBE} on R@100 by margins of 8.0\%, 8.9\%, and 4.4\%, respectively. Notably, compared to the biased backbone of MotifsNet, our method only sees a marginal decrease of 1.7\% on the R@K metric, whereas TransRwt, PPDL, and TBE experience more significant drops of 9.7\%, 10.6\%, and 6.1\% respectively against the same backbone. This underscores our method's capability to maintain high performance in head categories while aiming for debiasing predictions. Significantly, as shown in Table~\ref{OpenImagesV6}, our method achieved state-of-the-art performance on both mR@50 and R@50 metrics within the larger-scale dataset OI V6. This further demonstrates that the CAModule does not compromise the performance of head categories in pursuit of unbiased predictions.

\textbf{3)} Table~\ref{MRK} shows that our approach continues to maintain advantages on the composite metric MR@K. For instance, in the SGDet mode with the Transformer \cite{transformer} as the backbone, compared to the TransRwt \cite{TransRwt}, CFA \cite{CFA}, and DKBL \cite{DKBL} methods, our approach leads by 3.4\%, 2.7\%, and 2.2\% on ${\text{MR@K}}^{\Diamond}$, respectively. Moreover, our method significantly outperforms the biased backbones; for example, it achieves improvements of 8.2\%, 5.2\%, and 4.4\% over the Transformer backbone in the PredCls, SGCls, and SGDet modes, respectively. This superiority stems from the biased backbones' high R@K but very low mR@K, putting them at a disadvantage when computing MR@K. These results indicate that our method achieves a better balance between mR@K and R@K, enhancing the prediction accuracy for tail relationships while preserving the performance on head relationships.

\textit {Qualitative analyses.} In Fig. \ref{Qualitative}, we present some qualitative comparisons of our method against baseline models, from which we can observe:

\textbf{1)} Our approach demonstrates enhanced capability in recognizing fine-grained tail relationships, such as $\{$ \textit {$<$man, sitting on, chair$>$ vs $<$man, on, chair$>$} $\}$, and $\{$ \textit {$<$dog, laying on, bench$>$ vs $<$dog, on, bench$>$} $\}$. This improvement stems from our method's fine-tuned adjustments to the logits produced by baseline methods, where the adjustment factors are tail-friendly. This tail-friendliness is achieved by modeling these factors that consider the skewed distribution in the training data, specifically object, object pair,  co-occurrence, and relationship distributions.

\textbf{2)} Our method exhibits a stronger ability to capture spatial relationships between objects, for example, 
$\{$ \textit {$<$car1, behind, bench$>$ vs $<$car1, near, bench$>$} $\}$, and $\{$ \textit {$<$dog, in front of, person1$>$ vs $<$dog, near, person1$>$} $\}$. This benefit again arises from our enhanced recognition of tail relationships. Generally, when annotating relationships, people tend to label coarse relationships (such as \textit {near}) while overlooking more specific ones (like \textit {in front of}, \textit {behind}, \etc).

\textbf{3)} Compared to relatively easy spatial relationships, our method tends to predict more challenging semantic relationships, such as $\{$ \textit {$<$cat, watching, laptop$>$ vs $<$cat, near, laptop$>$} $\}$. We attribute this to the optimization of object pair distribution $\mathcal{P}$, where the incorporation of language model features allows the model to learn more semantic associations between objects.

\begin{table*}[htbp]
  \centering
  \caption{Results with different combinations of Euclidean distance $d$ and cosine similarity $cos$.}
    \resizebox{19.5cm}{!}{\begin{tabular}{ccccccccc}
    \midrule
    \multicolumn{1}{c|}{} &       & \multicolumn{1}{c|}{} & \multicolumn{2}{c|}{PredCls} & \multicolumn{2}{c|}{SGCls} & \multicolumn{2}{c}{SGDet} \\
    \multicolumn{1}{c|}{} & \multicolumn{1}{c}{$d$} & \multicolumn{1}{c|}{$cos$} & \multicolumn{1}{c|}{zR@20/50/100} & \multicolumn{1}{c|}{mR@20/50/100} & \multicolumn{1}{c|}{zR@20/50/100} & \multicolumn{1}{c|}{mR@20/50/100} & \multicolumn{1}{c|}{zR@20/50/100} & \multicolumn{1}{c}{mR@20/50/100} \\
    \midrule
    \multicolumn{1}{c|}{\multirow{3}[2]{*}{\rotatebox{90}{\scriptsize MotifsNet}}} & \ding{54}     & \multicolumn{1}{c|}{\ding{54}} & \multicolumn{1}{c|}{$7.0/10.8/13.5$} & \multicolumn{1}{c|}{$21.5/31.2/34.5$}     & \multicolumn{1}{c|}{$2.1/3.4/4.9$} & \multicolumn{1}{c|}{$14.2/17.1/19.8$}    & \multicolumn{1}{c|}{$0.3/1.1/2.5$} & \multicolumn{1}{c}{$10.2/14.1/15.8$} \\
    \multicolumn{1}{c|}{} & \multicolumn{1}{c}{\ding{54}}     & \multicolumn{1}{c|}{\ding{51}} & \multicolumn{1}{c|}{$8.9/12.5/16.3$} & \multicolumn{1}{c|}{$30.6/35.4/37.9$}     & \multicolumn{1}{c|}{$2.9/5.0/6.6$} & \multicolumn{1}{c|}{$17.4/19.5/22.8$}    & \multicolumn{1}{c|}{$1.1/2.6/3.4$} & \multicolumn{1}{c}{$11.3/15.2/17.5$} \\
    \multicolumn{1}{c|}{} & \multicolumn{1}{c}{\ding{51}}     & \multicolumn{1}{c|}{\ding{54}} & \multicolumn{1}{c|}{$8.4/11.6/15.7$} & \multicolumn{1}{c|}{$29.7/34.7/36.6$}     & \multicolumn{1}{c|}{$2.2/4.7/5.4$} & \multicolumn{1}{c|}{$17.2/18.8/21.7$}     & \multicolumn{1}{c|}{$0.8/2.6/3.1$} & \multicolumn{1}{c}{$10.2/14.9/17.1$} \\
    \multicolumn{1}{c|}{} & \multicolumn{1}{c}{\ding{51}}     & \multicolumn{1}{c|}{\ding{51}} & \multicolumn{1}{c|}{$9.5/16.7/20.3$} & \multicolumn{1}{c|}{$32.5/36.7/39.3$}     & \multicolumn{1}{c|}{$2.1/5.1/7.6$} & \multicolumn{1}{c|}{$18.8/21.1/24.7$}     & \multicolumn{1}{c|}{$1.4/2.1/3.7$} & \multicolumn{1}{c}{$11.7/16.3/18.2$} \\
    \midrule
    
    \multicolumn{1}{c|}{\multirow{3}[2]{*}{\rotatebox{90}{\scriptsize VCTree}}} & \ding{54}     & \multicolumn{1}{c|}{\ding{54}} & \multicolumn{1}{c|}{$6.6/11.3/13.8$} & \multicolumn{1}{c|}{$25.8/31.5/34.7$}     & \multicolumn{1}{c|}{$2.0/3.4/4.7$} & \multicolumn{1}{c|}{$13.7/17.2/20.3$}    & \multicolumn{1}{c|}{$0.3/0.9/2.3$} & \multicolumn{1}{c}{$9.4/14.3/16.1$} \\
    \multicolumn{1}{c|}{} & \multicolumn{1}{c}{\ding{54}}     & \multicolumn{1}{c|}{\ding{51}} & \multicolumn{1}{c|}{$8.2/12.9/16.6$} & \multicolumn{1}{c|}{$31.0/36.2/38.5$}     & \multicolumn{1}{c|}{$3.5/5.1/6.5$} & \multicolumn{1}{c|}{$18.1/20.8/24.1$}     & \multicolumn{1}{c|}{$0.9/2.1/2.7$} & \multicolumn{1}{c}{$11.4/15.8/18.3$} \\
    \multicolumn{1}{c|}{} & \multicolumn{1}{c}{\ding{51}}     & \multicolumn{1}{c|}{\ding{54}} & \multicolumn{1}{c|}{$7.9/11.3/16.1$} & \multicolumn{1}{c|}{$30.5/34.6/37.1$}     & \multicolumn{1}{c|}{$3.1/4.5/5.2$} & \multicolumn{1}{c|}{$18.4/20.4/23.3$}     & \multicolumn{1}{c|}{$0.8/1.7/2.4$} & \multicolumn{1}{c}{$11.0/15.2/17.6$} \\
    \multicolumn{1}{c|}{} & \multicolumn{1}{c}{\ding{51}}     & \multicolumn{1}{c|}{\ding{51}} & \multicolumn{1}{c|}{$8.1/14.5/18.4$} & \multicolumn{1}{c|}{$35.2/38.4/40.5$}     & \multicolumn{1}{c|}{$1.3/5.1/7.2$} & \multicolumn{1}{c|}{$21.7/23.2/25.9$}     & \multicolumn{1}{c|}{$1.1/1.6/3.2$} & \multicolumn{1}{c}{$12.3/16.4/19.6$} \\
    \midrule
    
    \multicolumn{1}{c|}{\multirow{3}[2]{*}{\rotatebox{90}{\scriptsize Transformer}}} & \ding{54}     & \multicolumn{1}{c|}{\ding{54}} & \multicolumn{1}{c|}{$6.4/9.6/13.1$} & \multicolumn{1}{c|}{$25.5/30.8/34.1$}     & \multicolumn{1}{c|}{$1.9/3.2/4.2$} & \multicolumn{1}{c|}{$13.8/16.7/19.6$}    & \multicolumn{1}{c|}{$0.2/0.7/2.1$} & \multicolumn{1}{c}{$9.8/14.0/15.5$} \\
    \multicolumn{1}{c|}{} & \multicolumn{1}{c}{\ding{54}}     & \multicolumn{1}{c|}{\ding{51}} & \multicolumn{1}{c|}{$7.7/11.3/15.5$} & \multicolumn{1}{c|}{$30.8/34.5/37.4$}     & \multicolumn{1}{c|}{$2.8/4.7/6.1$} & \multicolumn{1}{c|}{$17.0/19.2/22.4$}     & \multicolumn{1}{c|}{$1.1/2.3/3.2$} & \multicolumn{1}{c}{$11.9/16.3/18.4$} \\
    \multicolumn{1}{c|}{} & \multicolumn{1}{c}{\ding{51}}     & \multicolumn{1}{c|}{\ding{54}} & \multicolumn{1}{c|}{$6.3/10.6/15.1$} & \multicolumn{1}{c|}{$29.5/34.0/36.2$}     & \multicolumn{1}{c|}{$2.6/4.3/5.5$} & \multicolumn{1}{c|}{$16.9/18.5/21.8$}     & \multicolumn{1}{c|}{$0.9/1.7/2.8$} & \multicolumn{1}{c}{$11.3/15.5/17.9$} \\
    \multicolumn{1}{c|}{} & \multicolumn{1}{c}{\ding{51}}     & \multicolumn{1}{c|}{\ding{51}} & \multicolumn{1}{c|}{$8.5/15.4/19.7$} & \multicolumn{1}{c|}{$34.1/37.6/39.4$}     & \multicolumn{1}{c|}{$1.5/4.8/8.4$} & \multicolumn{1}{c|}{$16.7/21.3/23.9$}     & \multicolumn{1}{c|}{$1.2/1.7/4.6$} & \multicolumn{1}{c}{$12.2/16.6/19.3$} \\
    \midrule
    \end{tabular}}%
  \label{d-cos}%
\end{table*}%

\begin{table}[htbp]
  \centering
  \caption{Model performance with different logit adjustment methods.}
    \resizebox{8.5cm}{!}{\begin{tabular}{c|c|cc}
    \toprule
    \multicolumn{1}{r}{} &       & \multicolumn{1}{c}{CAModule} & \multicolumn{1}{c}{Vanilla logit adjustment} \\
        \multicolumn{1}{r}{} &       & \multicolumn{1}{c}{mR20/50/100} & \multicolumn{1}{c}{mR20/50/100} \\
    \midrule
    \multirow{3}[1]{*}{\rotatebox{90}{\tiny  MotifsNet}} &   PredCls     &   $32.5 / 36.7 / 39.3$    & $12.2/15.4 /16.7$ \\
          &  SGCls     & $18.8 /21.1 /24.7$      & $6.4 /7.6 /8.3$ \\
          &  SGDet     & $11.7 /16.3 /18.2$      &  $4.5 /5.9 /7.7$\\
    \midrule
    \multirow{3}[2]{*}{\rotatebox{90}{\tiny VCTree}} &   PredCls    &  $35.2 /38.4 /40.5$      &  $11.7 /15.2 /16.3$\\
          & SGCls      &  $21.7 /23.2 /25.9$     & $6.4 /7.6 /8.1$ \\
          &   SGDet    & $12.3 /16.4 /19.6$      &  $4.1 /5.5 /7.1$\\
    \midrule
    \multirow{3}[2]{*}{\rotatebox{90}{\tiny Transformer}} &  PredCls     & $34.1 /37.6 /39.4$      & $13.5 /16.6 /18.9$ \\
          &    SGCls   &   $16.7 /21.3 /23.9$    & $6.7 /8.6 /9.8$ \\
          &   SGDet    &   $12.2 /16.6 /19.3$    & $6.6 /8.3 /9.4$ \\
    \bottomrule
    \end{tabular}}%
  \label{logit-comparison}%
\end{table}%

\subsection{Ablations and Hyper-parameters}
\label{Ablations}

\textit {Compared to the vanilla logit adjustment method.} In this paper, we propose a fine-grained logit adjustment method, where each triplet corresponds to an adjustment factor, denoted by $\widetilde{\mathcal{A}}$ in Equation (\ref{triplet-level-adjustment}). Our method is inspired by logit adjustment techniques widely used in classification tasks, where typically each category is associated with an adjustment factor, represented by $\mathcal{A}$ in Equation (\ref{relation-level-adjust}). In Section \ref{sec3.2}, we provide a detailed analysis of why this vanilla logit adjustment method is not suitable for the SGG task, fundamentally due to the long-tail distribution problem still present in triplets formed by the same relationship. Here, we further empirically compare with the vanilla logit adjustment method, referencing a common practice where $\mathcal{A}$ is set to the inverse of the category frequency: 
\[
    \mathcal{A}=\{\frac{1}{\mathcal{R}_1}, \frac{1}{\mathcal{R}_2}, \ldots, \frac{1}{\mathcal{R}_{N_r}}\}.
\]
Consistent with Equation (\ref{R-distribution}), $\mathcal{R}_i$ denotes the count of the $i$-th relationship category. The comparison results, presented in Table~\ref{logit-comparison}, reveal that the vanilla logit adjustment method has almost no effect in adjusting the bias issue of the SGG model, thereby reinforceing our initial analysis.

\textit {Optimizing $\mathcal{P}$ for enhancing zero-shot capabilities.}
In this paper, we introduce the CAModule, which calculates a set of fine-grained logit adjustment factors based on distribution knowledge extracted from observed data $\mathcal{D}$, for adjusting biased SGG models. As demonstrated in Section \ref{results}, our approach achieves state-of-the-art performance on the debiasing metric mR@K. Surprisingly, we also observe significant advancements on the challenging zero-shot metric zR@K. As indicated by the results labeled as ``$\text{CAModule}^{-\mathcal{P}}$'' in Table~\ref{zRk}, which represent outcomes without optimizing the object pair distribution $\mathcal{P}$. Under this vanilla setting, for instance, using the $\text{AVG}_{\text{zR}}^{\Diamond}$ metric with the PredCls evaluation mode, our method achieves notable improvements in the MotifsNet \cite{Neuralmotifs}, VCTree \cite{VCtree}, and Transformer backbones \cite{transformer}, enhancing performance by 6.4\%, 3.6\%, and 1.8\%, respectively. We attribute this improvement primarily to our proposed causal modeling approach, which enables the construction of some zero-shot relationships. However, further investigation reveals that this combination relies on existing object pairs in the observed data $\mathcal{D}$, meaning it can only generate zero-shot instances based on these pairs and is ineffective for zero-shot pairs (refer to the analysis in Section \ref{sec3.4}). To address this issue, we introduce two inference rules to speculate on potential zero-shot pairs and optimize the object pair distribution $\mathcal{P}$.

These rules function by identifying plausible zero-shot object pairs based on attribute similarity and integrating them into the object pair distribution $\mathcal{P}$. Specifically, for each inferred pair $<$ $o_m, o_M $ $>$ derived from Rule 1 or Rule 2, a weighting factor related to the attribute similarity $S_{o_m, o_M}$ (as defined in Section \ref{sec3.4}) is applied to optimize $\mathcal{P}$. This adjustment not only extends the coverage of $\mathcal{P}$ beyond observed pairs in $\mathcal{D}$, but also ensures that plausible zero-shot pairs inferred by the rules are appropriately weighted according to their similarity scores. Furthermore, the enhanced $\mathcal{P}$ better aligns with real-world plausibility while maintaining consistency with the underlying causal structure. As shown in Table~\ref{zRk}, in the PredCls mode, compared to the vanilla setting $\text{CAModule}^{-\mathcal{P}}$, the optimized approach shows enhancements of 6.3\%, 3.9\%, and 6.2\% across three backbones respectively, and compared to the original backbone setting, the increases are 12.7\%, 7.5\%, and 8.0\%. These improvements validate the effectiveness of combining Rule 1 and Rule 2 with $\mathcal{P}$. By leveraging attribute similarity to identify and weight zero-shot pairs, the enhanced $\mathcal{P}$ distribution significantly contributes to the model’s capability to infer unseen relationships. Thus, our rules-based optimization approach effectively complements the existing causal modeling framework, resulting in a more comprehensive and robust zero-shot relationship recognition.

\begin{figure}
    \footnotesize\centering
    \centerline{\includegraphics[width=0.95\linewidth]{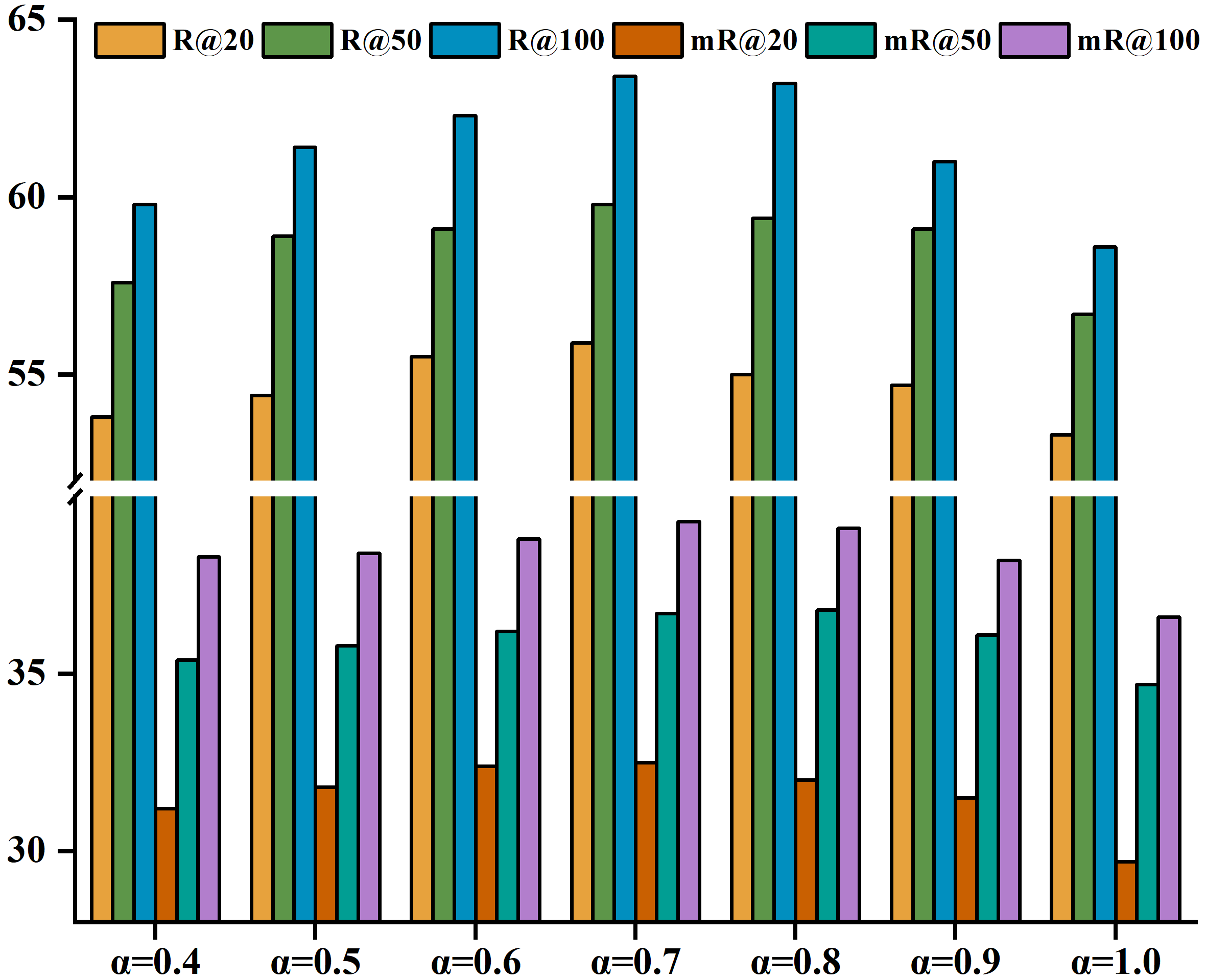}}
        \vspace{-0.2cm}
        \caption{Model performance with different $\alpha$. The baseline SGG model and evaluation mode used here are MotifsNet \cite{Neuralmotifs} and PredCls, respectively.}
        \label{alpha}
\end{figure}

\begin{figure}
    \footnotesize\centering
    \centerline{\includegraphics[width=0.95\linewidth]{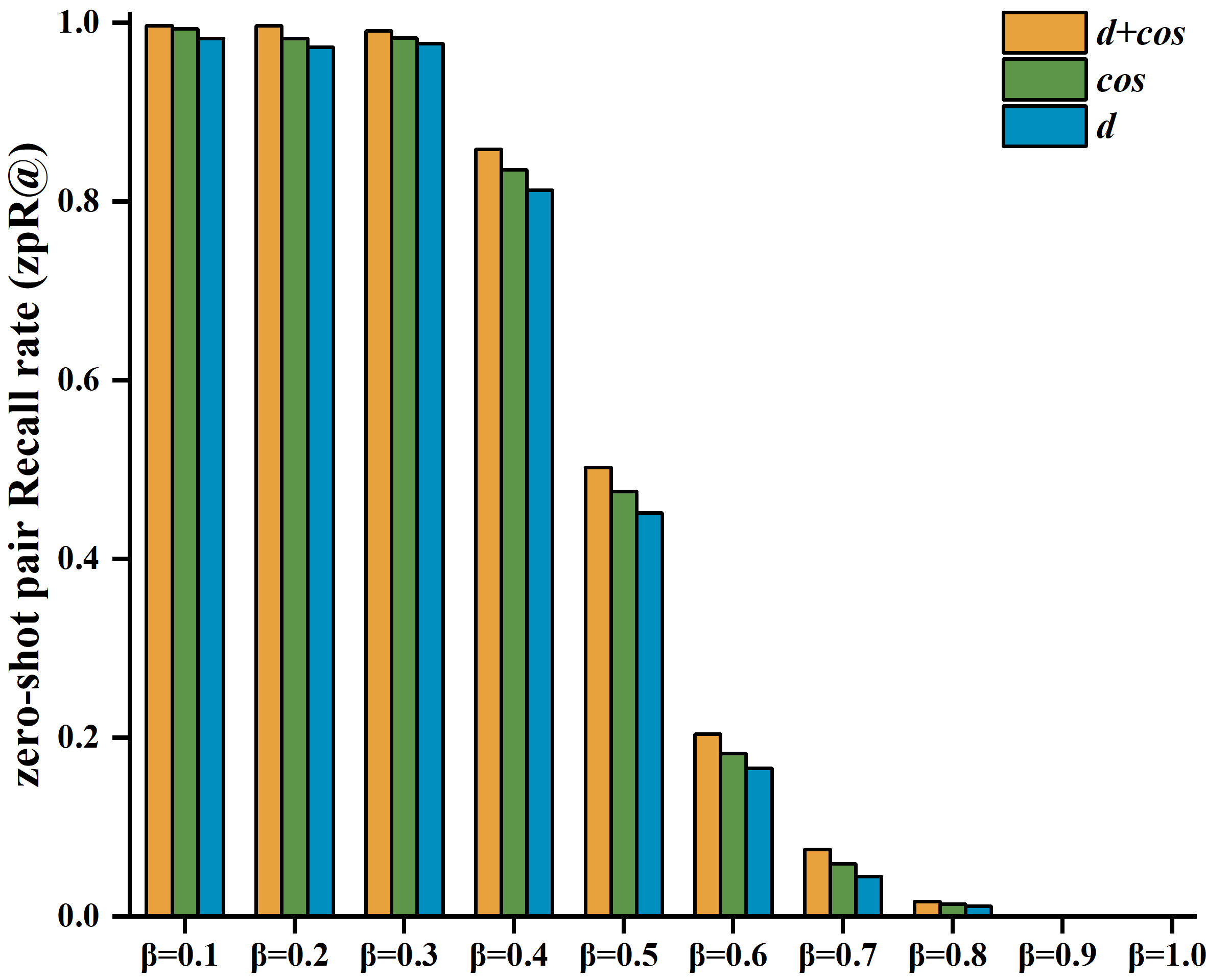}}
        \vspace{-0.2cm}
        \caption{Zero-shot object pair Recall rate (zpR@) with different $\beta$. The baseline SGG model and evaluation mode used here are MotifsNet \cite{Neuralmotifs} and PredCls, respectively.}
        \label{beta}
\end{figure}

\textit{Euclidean distance and cosine similarity.} In Section \ref{sec3.4}), we introduce \textbf{Rule 1} and \textbf{Rule 2} to infer potential zero-shot pairs, further optimizing the object pair distribution $\mathcal{P}$ by assessing the similarity of attributes between two object categories, derived from the similarity of their features. However, we observe that for the similarity of object attributes in the SGG task, both the absolute distance and direction of object features are critically important, with detailed analysis provided in Section \ref{sec3.4}). In response, we employ both Euclidean distance and cosine similarity concurrently to capture similarities in distance and direction, as outlined in Equation (\ref{distance-calculation}). Ablation studies of these two similarity calculation methods, presented in Table~\ref{d-cos}, indicate: \textbf{1)} Both Euclidean distance and cosine similarity enhance model performance, including on the challenging zero-shot metric zR@K and the debiasing metric mR@K; \textbf{2)} Combining Euclidean distance and cosine similarity yields better results than using either alone; \textbf{3)} Cosine similarity outperforms Euclidean distance. For instance, under the PredCls evaluation mode with the MotifsNet backbone, Euclidean distance contributes a 2.1\% improvement over the unoptimized $\mathcal{P}$ baseline, whereas for cosine similarity, it's 3.4\%. This demonstrates that directionality plays a more significant role than distance in assessing object attribute similarity for the SGG task.

\textit{The hyper-parameter $\alpha$.} In Equation (\ref{distance-calculation}), $\alpha$ is utilized to weight the two similarity calculation methods, where a higher value emphasizes cosine similarity, while a lower value highlights Euclidean distance. Equation (\ref{distance-calculation}) employs a hybrid similarity computation approach that not only highlights the importance of object attributes (Euclidean distance) but also emphasizes the significance of object directions (cosine similarity). Specifically, we consider the directionality of objects to be crucial for SGG tasks, especially since \textit {$<$girl, chair$>$ } and \textit {$<$chair, girl$>$ } represent two entirely distinct object pairs. We experimented with different values of $\alpha$ from 0.4 to 1.0, in increments of 0.1, as depicted in  Fig. \ref{alpha}. The results indicate: \textbf{1)} The model performs optimally when $\alpha$ is set to 0.7. \textbf{2)} Performance continuously declines as $\alpha$ exceeds 0.7, suggesting that an excessive emphasis on cosine similarity, at the expense of capturing differences in magnitude, is detrimental. Similarly, a decrease in $\alpha$ below 0.7 also results in a performance decline, attributed to the gradual loss of directional similarity. \textbf{3)} Both cosine similarity and Euclidean distance contribute to the calculation of object attribute similarity, with cosine similarity playing a more crucial role. This observation aligns with our findings presented in Table~\ref{d-cos}.

\begin{figure}
    \footnotesize\centering
    \centerline{\includegraphics[width=0.95\linewidth]{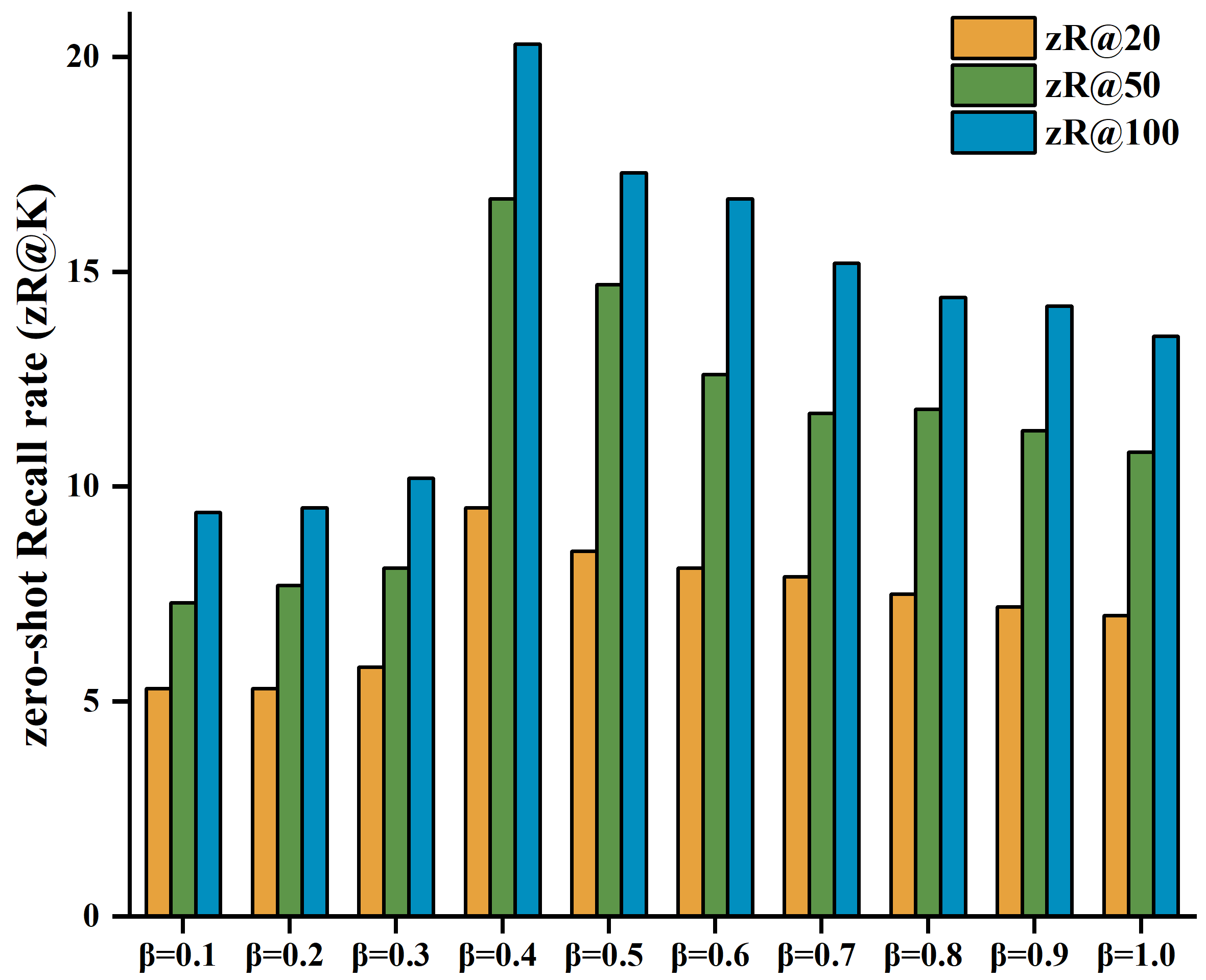}}
        \vspace{-0.2cm}
        \caption{Zero-shot performance (z@K) with different $\beta$. The baseline SGG model and evaluation mode used here are MotifsNet \cite{Neuralmotifs} and PredCls, respectively.}
        \label{different_beta}
\end{figure}

\textit{The hyper-parameter $\beta$.} $\beta$ serves as a threshold for determining the similarity of object attributes. If the distance between two objects (calculated using Equation (\ref{distance-calculation})) exceeds $\beta$, they are deemed to have similar attributes; otherwise, they are considered dissimilar. In Fig. \ref{beta}, we introduce a new metric, the recall rate for zero-shot object pairs (zpR@), representing the proportion of successfully inferred zero-shot pairs out of the total number of zero-shot pairs. It is important to note that zpR@ should be interpreted with caution because, due to dataset limitations, some inherently plausible object pairs might not be annotated. In such cases, our method's inference of these relationships, while appearing reasonable, will not increase zpR@. Nevertheless, this does not detract from using zpR@ to validate the effectiveness of our approach in inferring zero-shot object pairs. The results in Fig. \ref{beta} yield the following observations: \textbf{1)} At low values, \eg, $\beta=0.1, 0.2, 0.3$, zpR@ approaches nearly 100\%. However, these values are not necessarily optimal as a low $\beta$ makes the criterion for attribute similarity too lenient, potentially leading to the inference of many implausible object pairs; \textbf{2)} At high values, particularly when $\beta$ equals 0.9 or 1.0, zpR@ drops to 0\%. This occurs because a high threshold imposes a stringent criterion for attribute similarity, making it challenging to infer even object pairs with similar attributes. Further, Fig. \ref{beta} suggests specific hypotheses: When $\beta$ is set to 0.1, 0.2, or 0.3, the zpR@ values are quite similar, indicating that these thresholds may be too loose and classify some impossible object pairs as zero-shot pairs; however, as $\beta$ increases from 0.4 to 0.5, there is a precipitous drop in zpR@ values, suggesting that setting $\beta$ above 0.4 might omit some zero-shot pairs. Thus, we set $\beta$ at 0.4 in this paper. To validate our hypothesis, we evaluated zero-shot performance at different $\beta$ settings in Fig. \ref{different_beta}. We observed that zR@K is strongly influenced by zpR@, particularly at low (\eg, $\beta$ = 0.1, 0.2, 0.3) or high values (\eg, $\beta$ = 0.8, 0.9, 1.0), both of which yielded unsatisfactory zR@K. The strongest zero-shot capability was at $\beta = 0.4$, consistent with findings in Fig. \ref{beta} that this $\beta$ value provided the best zpR@. This evidence confirms our hypothesis and supports our method of choosing the optimal $\beta$ based on zpR@. Interestingly, even at $\beta = 1.0$, where no potential zero-shot pair is captured, our method still outperforms the baseline framework, further demonstrating our proposed causal structure's ability to uncover potential zero-shot pairs.

\begin{table}[t]
    \centering
    \caption{Model inference time of CAModule and baseline methods on VG150 dataset. The time reported here is s/img per device.}
        \begin{tabular}{l|ccc}
        \toprule
              & \multicolumn{2}{c}{PredCls} \\
              & mR@20/50/100  & testing  & training   \\
        \midrule
        \midrule
        MotifsNet (backbone) \cite{Neuralmotifs}  & $12.2 / 15.5 / 16.8$ & $0.11825$ & $0.11943$ \\
        \quad \textbf{CAModule (ours)} &$32.5/36.7/39.3$ & $0.12382$ & $0.12517$     \\
        \midrule
        VCTree (backbone) \cite{VCtree}  & $12.4 / 15.4 / 16.6$ & $0.15380$ & $0.15526$ \\
        \quad \textbf{CAModule (ours)} & $35.2/38.4 / 40.5$& $0.15983$   & $0.16037$ \\
        \midrule
        Transformer (backbone) \cite{transformer}  & $12.4/16.0/17.5$& $0.18456$ & $0.18522$ \\
        \quad \textbf{CAModule (ours)} &$34.1 / 37.6 / 39.4$ & $0.19231$   & $0.19299$   \\
        \bottomrule
        \end{tabular}%
    \label{inference_time}%
    \vspace{-0.2cm}
\end{table}

\textit{Model efficiency analysis.} In this paper, we propose a causal adjustment module, \ie, CAModule, which can be flexibly integrated into any existing SGG framework. While it effectively mitigates model biases in the baseline framework, it inevitably introduces additional computational overhead. However, the CAModule is optimized to be lightweight, ensuring minimal additional computational cost by:

\textbf{1)} Employing low-dimensional statistics from the training set ($\mathcal{O}$, $\mathcal{C}$, $\mathcal{P}$, $\mathcal{R}$) for input, simplifying the feature space necessary for causal analysis and thereby curbing the complexity.

\textbf{2)} Using just three transformer blocks arranged in a cascade to model the causal interactions among $\mathcal{O}$, $\mathcal{C}$, $\mathcal{P}$, and $\mathcal{R}$, as shown in Fig. \ref{CalculatingAdjustmentFactors}. We have intentionally constrained the complexity within the transformers, with embedding dimensions for $\mathcal{O}$, $\mathcal{C}$, $\mathcal{P}$, and $\mathcal{R}$ limited to 128, as detailed in Algorithm \ref{algorithm1}.

We report the typical SGG frameworks and the model inference time after integrating the CAModule in Table~\ref{inference_time}. For an objective comparison of efficiency, both testing and training were conducted on an RTX 3090 using CUDA 11.3 and PyTorch 1.10.1. The results in Table~\ref{inference_time} indicate that integrating the CAModule increases both model inference and training times by approximately 3\% to 5\%; however, it significantly enhances the model's performance on unbiased metrics (mR@K). Therefore, we argue that the additional inference and training cost introduced by the CAModule is negligible. Nonetheless, researching more efficient modules to capture causal relationships in skewed distributions within SGG datasets remains an area for future work.

\section{Conclusion}
\label{section5}
In this paper, we introduce the Causal Adjustment Module (CAModule), which inputs causal variables from the Mediator-based Causal Chain Model (MCCM) to output a set of triplet-level adjustment factors aimed at correcting biased predictions in Scene Graph Generation (SGG) models. MCCM models the causality among objects, object pairs, and relationships, highlighting the skewed distributions of objects and object pairs as deeper sources of bias in models, aspects frequently neglected in existing studies. As a key improvement, MCCM incorporates mediator variables to further enrich the causality between objects and object pairs, thereby enhancing the debiasing capability. Moreover, MCCM's ability to compose zero-shot relationships enhances the model's recognition capabilities for such relationships, with our optimization of the object pair distribution further enhancing this capacity. CAModule achieves state-of-the-art performance on the critical debiasing metric, mean recall rate, and also demonstrates substantial progress on the challenging zero-shot recall rate metric. 

The CAModule's advancement in recognizing zero-shot relationships is encouraging. In the future,  we will continue refining our approach to further enhance this capability, which we believe is crucial for achieving complete debiasing in SGG.

\ifCLASSOPTIONcaptionsoff
  \newpage
\fi

\bibliographystyle{IEEEtran}
\bibliography{biblio}

\end{document}